\documentclass{article} 
\usepackage{iclr2027_conference,times}
\renewcommand{\iclrruler}[1]{}

\usepackage{amsmath,amsfonts,bm}

\def\eqref#1{equation~\ref{#1}}
\def\Eqref#1{Equation~\ref{#1}}

\def\1{\bm{1}}

\DeclareMathAlphabet{\mathsfit}{\encodingdefault}{\sfdefault}{m}{sl}
\SetMathAlphabet{\mathsfit}{bold}{\encodingdefault}{\sfdefault}{bx}{n}

\usepackage{hyperref}
\usepackage{url}

\usepackage{mathtools}
\usepackage{amsmath}
\usepackage{booktabs}
\usepackage{multirow}
\usepackage{graphicx}
\def\algo{{\textsc{RSP}}}
\usepackage{amssymb}
\usepackage{adjustbox}

\usepackage{colortbl}
\usepackage{makecell}

\definecolor{dropSmall}{HTML}{D97706}
\definecolor{dropMedium}{HTML}{EA580C}
\definecolor{dropLarge}{HTML}{DC2626}
\definecolor{dropSevere}{HTML}{991B1B}
\definecolor{oursBG}{HTML}{F0FDF4}

\newcommand{\dcell}[3]{%
#1\,{\scriptsize\textcolor{#3}{(#2)}}%
}

\usepackage[most]{tcolorbox}
\usepackage{xcolor}
\usepackage{amsmath,amssymb}

\definecolor{ExampleBG}{RGB}{247,247,247}
\definecolor{ExampleTitleBG}{RGB}{220,220,220}
\definecolor{ExampleFrame}{RGB}{195,195,195}

\definecolor{CorrectColor}{RGB}{62,115,78}
\definecolor{IncorrectColor}{RGB}{160,65,65}

\definecolor{BreakColor}{RGB}{145,100,65}
\definecolor{RepairColor}{RGB}{70,100,135}

\newtcolorbox{prmexample}[1]{
    enhanced,
    breakable,
    colback=ExampleBG,
    colframe=ExampleFrame,
    boxrule=0.5pt,
    arc=1.2mm,
    left=6mm,
    right=6mm,
    top=4mm,
    bottom=4mm,
    title={#1},
    colbacktitle=ExampleTitleBG,
    coltitle=black,
    fonttitle=\bfseries,
    before skip=8pt,
    after skip=10pt
}

\usepackage{siunitx}

\definecolor{CorrectColor}{RGB}{62,115,78}
\definecolor{IncorrectColor}{RGB}{160,65,65}

\definecolor{BreakColor}{RGB}{145,100,65}
\definecolor{RepairColor}{RGB}{70,100,135}

\newcommand{\correctmark}{%
    \textcolor{CorrectColor}{$\checkmark$}%
}

\newcommand{\incorrectmark}{%
    \textcolor{IncorrectColor}{$\times$}%
}

\newcommand{\scorethree}[1]{%
    \num[
        round-mode=places,
        round-precision=3,
        minimum-decimal-digits=3
    ]{#1}%
}

\newcommand{\correctmetrics}[3]{%
    {\footnotesize
    \correctmark
    \hspace{0.35em}
    \textcolor{CorrectColor}{%
        \textbf{PRM}~\scorethree{#1}%
    }%
    \hspace{0.8em}
    \textcolor{BreakColor}{%
        \textbf{Break}~\scorethree{#2}%
    }%
    \hspace{0.8em}
    \textcolor{RepairColor}{%
        \textbf{Repair}~\scorethree{#3}%
    }%
    }%
}

\newcommand{\incorrectmetrics}[3]{%
    {\footnotesize
    \incorrectmark
    \hspace{0.35em}
    \textcolor{IncorrectColor}{%
        \textbf{PRM}~\scorethree{#1}%
    }%
    \hspace{0.8em}
    \textcolor{BreakColor}{%
        \textbf{Break}~\scorethree{#2}%
    }%
    \hspace{0.8em}
    \textcolor{RepairColor}{%
        \textbf{Repair}~\scorethree{#3}%
    }%
    }%
}

\newcommand{\CorrectStepHeader}[4]{%
    \par\vspace{0.55em}%
    \noindent
    \textbf{Step #1:}%
    \hfill
    \correctmetrics{#2}{#3}{#4}%
    \par\vspace{0.12em}%
}

\newcommand{\IncorrectStepHeader}[4]{%
    \par\vspace{0.55em}%
    \noindent
    \textbf{Step #1:}%
    \hfill
    \incorrectmetrics{#2}{#3}{#4}%
    \par\vspace{0.12em}%
}

\title{Learning Process Rewards via Reasoning State Propagation}

\author{
\textbf{Kai Gan}$^{1,2}$,
\textbf{Zi-Hao Zhou}$^{1,2}$,
\textbf{Bo Ye}$^{1,2,3}$,
\textbf{Jian Zhao}$^{3,4}$,
\textbf{Min-Ling Zhang}$^{1,2}$,
\textbf{Tong Wei}$^{1,2,\dagger}$
\\[3pt]
$^{1}$School of Computer Science and Engineering, Southeast University,
Nanjing 210096, China \\
$^{2}$Key Laboratory of Computer Network and Information Integration
(Southeast University), \\
\phantom{$^{2}$}Ministry of Education, China \\
$^{3}$Zhongguancun Academy \\
$^{4}$Zhongguancun Institute of Artificial Intelligence
}

\iclrfinalcopy 
\begin{document}

\maketitle

\begin{abstract}

Process reward models (PRMs) have demonstrated notable effectiveness in test-time scaling and reinforcement learning by providing fine-grained signals for evaluating intermediate reasoning states, but their training relies heavily on costly process annotations. A natural way to alleviate this dependence is to complement limited process supervision with scalable outcome supervision. However, existing PRMs often model reasoning prefixes independently, providing no explicit mechanism for effectively using final outcome to guide the learning of intermediate reasoning states.
We introduce \textit{Reasoning State Propagation (\algo)}, which represents each reasoning prefix with a binary validity state and models transitions between successive states across the reasoning trajectory. Specifically, \algo{} predicts a \emph{break} probability that a valid state becomes invalid and a \emph{repair} probability that an invalid state returns to valid. By propagating these transitions, \algo{} connects intermediate states to the final state, allowing process annotations to supervise intermediate states while outcome labels supervise the final state and can provide learning signals to preceding steps. Across reasoning search, response selection, and reinforcement learning, \algo{} consistently outperforms representative PRM baselines, with average improvements over Qwen2.5-Math-PRM of 5.6\% in beam search and 2.1\% in reinforcement learning.




\end{abstract}

\section{Introduction}
Reinforcement learning with verifiable rewards (RLVR) \citep{jaech2024openai,guo2025deepseek,team2025kimi,yu2026dapo} has driven substantial progress in large reasoning models (LRMs) \citep{yang2025qwen3,seed2025seed1,xiaomi2025mimo}. However, RLVR typically relies on final-answer correctness as a trajectory-level reward, reducing a multi-step reasoning process to a single scalar signal. Such coarse supervision cannot distinguish the contributions of individual reasoning steps, making fine-grained credit assignment difficult. This limitation is especially problematic for all-incorrect groups on challenging problems. In group-relative methods such as GRPO \citep{shao2024deepseekmath}, identical rewards within all-incorrect groups result in zero advantages after group normalization, leaving no learning signal for policy optimization. To address these limitations, PRMs~\citep{lightman2024let,wang2024math,luo2024improve,zhang2024rest,li2025process,zhang2025lessons,yang2025treerpo,zhang2025linking} have emerged to provide step-level feedback throughout the reasoning process.

Despite the notable success of PRMs in test-time scaling \citep{lightman2024let,lu2024autopsv,snell2025scaling} and reinforcement learning \citep{li2026step,agrawal2026verigate}, PRM training typically relies on process annotations, which are costly and difficult to obtain at scale. Such annotations provide supervision over intermediate reasoning steps and are particularly valuable for fine-grained process evaluation. In contrast, outcome labels are substantially cheaper and readily scalable in verifiable domains since they only require determining whether the complete reasoning trajectory ultimately reaches a correct answer. This disparity suggests a natural complementary supervision strategy: scarce process annotations provide guidance on intermediate reasoning states, while abundant outcome labels offer scalable supervision over a much larger collection of trajectories. However, how to effectively integrate these two signals within a PRM remains underexplored.

\begin{figure*}[t]
    \centering

    \begin{adjustbox}{max width=\textwidth}
        \includegraphics[height=7.0cm]{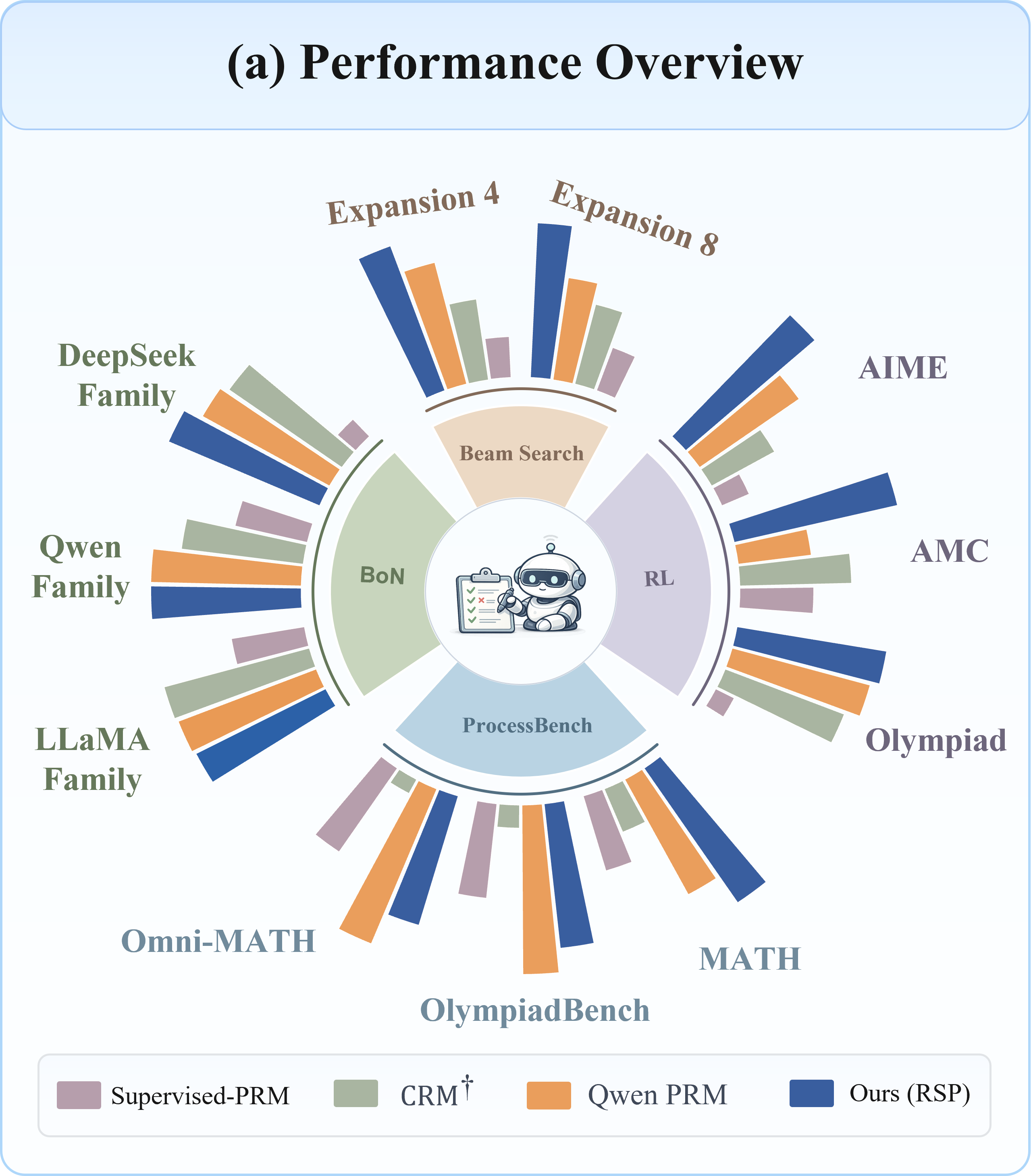}
        \hspace{0.1cm}
        \includegraphics[height=7.0cm]{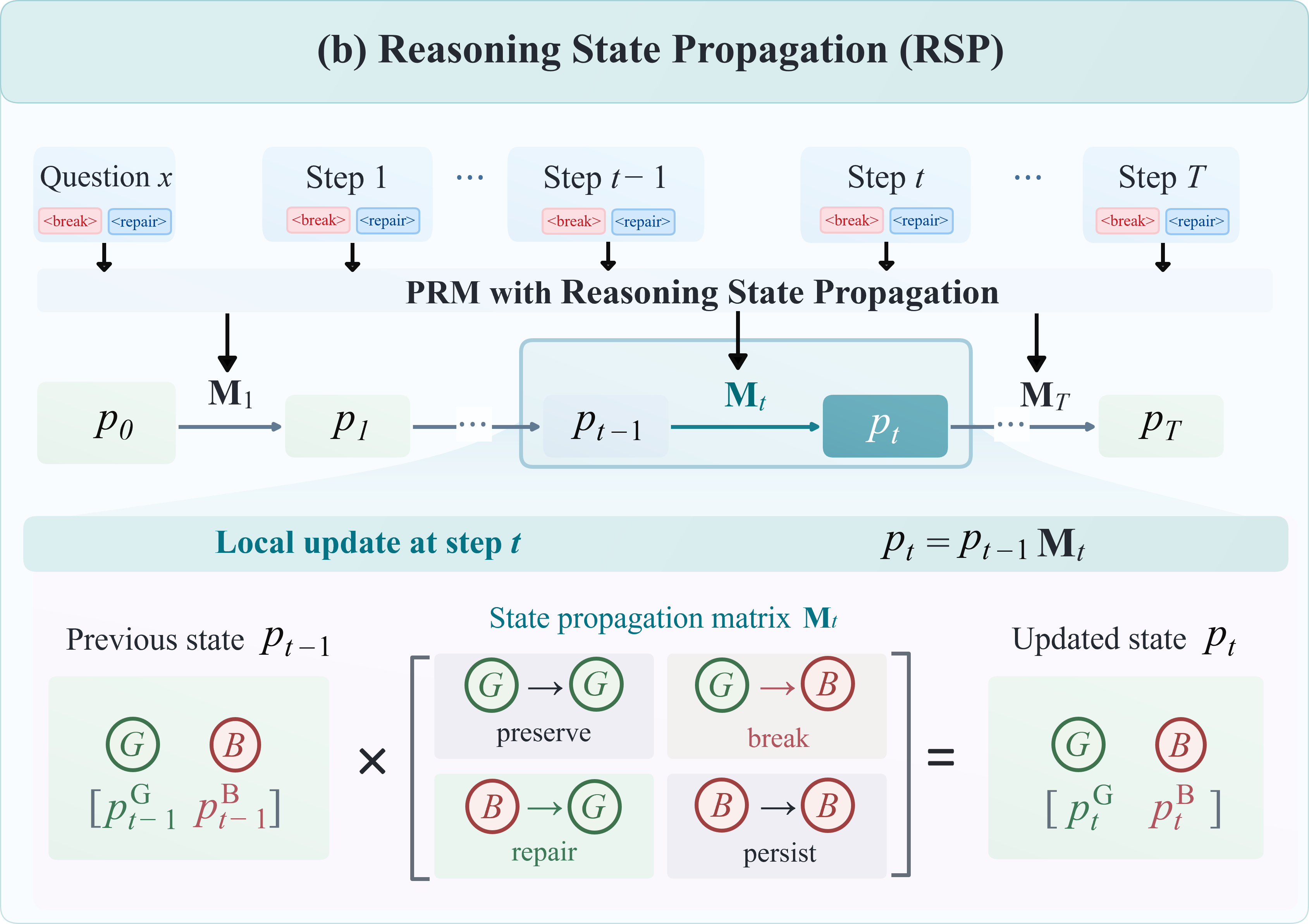}
    \end{adjustbox}

\caption{
(a) Empirical performance across evaluation settings.
(b) \algo{} propagates a binary reasoning state across steps using \emph{break} and \emph{repair} probabilities.
$\mathrm{G}$ denotes a \emph{Good or Valid} state with no unresolved error, while $\mathrm{B}$ denotes a \emph{Bad or Invalid} state with an unresolved error.
Transitions $\mathrm{G}\!\rightarrow\!\mathrm{B}$ and $\mathrm{B}\!\rightarrow\!\mathrm{G}$ correspond to error introduction (\emph{break}) and error recovery (\emph{repair}), respectively.
}
    \label{fig:overview}
\end{figure*}

A key obstacle to effectively exploiting this complementary supervision is how to relate intermediate reasoning states to the final outcome. Existing discriminative PRMs optimize individual prefixes as independent classification targets~\citep{lightman2024let,zhang2025lessons}, providing no explicit structure for connecting reasoning states across steps. Recent methods such as PQM~\citep{li2025process} and CRM~\citep{zhang2025linking} introduce sequential dependencies, but largely organize the reasoning process around the first erroneous step and effectively treat subsequent reasoning as remaining invalid. Such a formulation oversimplifies the relationship between intermediate reasoning states and final correctness. An intermediate error may be corrected by later reasoning and therefore does not necessarily lead to an incorrect final outcome. Similarly, a correct final outcome does not imply that every preceding reasoning step is valid. Another limitation is reflected in the popular process annotation protocol of PRM800K \citep{lightman2024let}, where annotation terminates once the first error is identified, leaving subsequent reasoning unannotated. In fact, modern LRMs can detect and revise earlier mistakes during continued reasoning~\citep{guo2025deepseek,lee2026reasoningflow}. This motivates a more general formulation that explicitly propagates reasoning states across steps.


In this paper, we introduce \textbf{Reasoning State Propagation (\algo)} for PRM training. Rather than independently predicting the validity of each reasoning prefix, \algo{} associates each prefix with an effective reasoning state that indicates whether the currently maintained derivation is valid or contains an unresolved error and propagates this state across successive prefixes. Consider a reasoning trajectory composed of steps $(s_1,\ldots,s_T)$, where $s_t$ denotes the $t$-th reasoning step. When extending the preceding prefix $\mathbf{s}_{<t}$ with step $s_t$, \algo{} estimates two conditional update probabilities: a \emph{break} probability, which measures the likelihood that a previously valid state becomes invalid after incorporating $s_t$, and a \emph{repair} probability, which measures the likelihood that the new step resolves or supersedes an earlier error and returns the current reasoning state to a valid state. These probabilities determine how the state of the preceding prefix is updated after incorporating $s_t$. Repeating this process throughout the trajectory explicitly links the reasoning states across steps. The framework of \algo{} is shown in Figure~\ref{fig:overview}(b).


\algo{} provides a natural interface for jointly leveraging process and outcome supervision. Process annotations can ground the validity of intermediate reasoning states, while outcome labels supervise the final propagated state of the complete trajectory. Since the final state depends on the sequence of preceding state updates, outcome supervision can supervise the final state and provide learning signals through the intervening state updates. This provides a way for outcome feedback to complement the model's locally grounded validity predictions on trajectories without process labels.

Our main contributions are summarized as follows:
\begin{itemize}
    \setlength{\itemsep}{5pt}
    \setlength{\parskip}{0pt}
    \setlength{\parsep}{0pt}
    \setlength{\topsep}{2pt}

    \item We propose a new formulation \algo{} for process reward modeling that models transitions between successive reasoning states through \emph{break} and \emph{repair}, rather than treating each reasoning prefix independently, thus explicitly linking reasoning states across the trajectory.



    
    \item \algo{} jointly leverages scarce process annotations and scalable outcome supervision, with process labels grounding intermediate states and outcome labels constraining the final state. We further analyze how both signals provide training signals for state updates.





    \item Extensive experiments across test-time scaling and reinforcement learning demonstrate consistent improvements over representative PRMs. Comprehensive analyses further validate the effectiveness of reasoning state propagation and its key design choices.


\end{itemize}

\section{Related Work}
\label{related_work}

\subsection{Enhancing Reasoning in LRMs}
Recent studies improve the reasoning capabilities of LRMs through test-time scaling and reinforcement learning. \citet{lightman2024let} use process reward models to rank sampled solutions in Best-of-$N$ selection. \citet{xie2023self} introduce self-evaluation-guided beam search, which prunes intermediate reasoning prefixes during generation. REST-MCTS* \citep{zhang2024rest} further incorporates process rewards into Monte Carlo search and uses the resulting trajectories for self-training. Beyond test-time scaling, DeepSeek-R1 \citep{guo2025deepseek} demonstrates that reinforcement learning with verifiable outcomes can substantially strengthen reasoning capabilities. PURE \citep{cheng2026stop} introduces min-form credit assignment, using future process rewards to provide denser supervision for intermediate reasoning steps. VeriGate \citep{agrawal2026verigate} selectively activates PRM-based token-level supervision when outcome rewards fail to provide group-relative learning signal. These advances increasingly rely on reliable and fine-grained process feedback, motivating accurate and scalable PRMs for evaluating intermediate reasoning states.


\subsection{Process Reward Modeling}
PRMs rely on fine-grained signals about intermediate reasoning steps, which are commonly obtained through human annotation~\citep{uesato2022solving,lightman2024let}, rollout-based estimation~\citep{wang2024math,tan2025enhancing}, or automated judging~\citep{zhang2025lessons,she2025r}. PRM800K~\citep{lightman2024let} collects human judgments of step-level correctness and trains discriminative PRMs to classify each reasoning step as valid or invalid. Math-Shepherd~\citep{wang2024math} reduces annotation costs by estimating step correctness from whether sampled continuations can reach the correct final answer. \citet{zhang2025lessons} improve automatically constructed labels through consensus filtering. With these labels, conventional PRMs typically optimize step-wise classification objectives without modeling dependencies between adjacent reasoning prefixes. OVM~\citep{yu2024ovm} learns the probability that a partial trajectory can eventually reach a correct answer using outcome-annotated rollouts. PQM~\citep{li2025process} formulates process reward modeling as Q-value ranking, capturing relative relationships among steps within a trajectory. CRM~\citep{zhang2025linking} models the first invalid state and links the process rewards to the final outcome using the probability chain rule. Other recent studies improve process reward modeling through dynamic evaluation criteria or retrieval-augmented verification
\citep{yin2025dynamic,zhu2025retrieval}. However, existing PRMs do not jointly exploit process- and outcome-level supervision and explicitly model flexible transitions between reasoning states.

\section{Method}
\label{method}

In this section, we present \algo{}, including its reasoning state propagation formulation, model parameterization, and learning objectives.

\subsection{Preliminaries}
\paragraph{Task Definition}
Given a question $x$, an LRM policy $\pi$ generates a reasoning trajectory $\mathbf{s}=(s_1,\ldots,s_T)$ autoregressively, where
$
s_t \sim \pi(\cdot \mid x,\mathbf{s}_{<t})
$
and $\mathbf{s}_{<t}=(s_1,\ldots,s_{t-1})$ denotes the preceding reasoning prefix. We consider two types of supervision. For process-annotated trajectories, binary process labels
$y_t^{\mathrm{step}} \in \{0,1\}$ are available for each step, where $y_t^{\mathrm{step}}=1$ indicates that the reasoning prefix
$\mathbf{s}_{\leq t}$ is valid after incorporating step $s_t$. For outcome-annotated trajectories, only a final label $y^{\mathrm{out}} \in \{0,1\}$ is available, indicating whether the complete trajectory reaches a correct answer. Let $N_{\mathrm{step}}$ and $N_{\mathrm{out}}$ denote the numbers of
process-annotated and outcome-annotated trajectories. For the $i$-th trajectory, $T_i$ denotes its number of reasoning steps, and
$\mathcal{A}_i \subseteq \{1,\ldots,T_i\}$ denotes the set of steps with available process annotations. The two training sets are defined as
$
\mathcal{D}_{\mathrm{step}}
=
\left\{
\left(
x_i,
\mathbf{s}_i,
\{y_{i,t}^{\mathrm{step}}\}_{t\in\mathcal{A}_i}
\right)
\right\}_{i=1}^{N_{\mathrm{step}}},
$
and
$
\mathcal{D}_{\mathrm{out}}
=
\left\{
\left(
x_i,
\mathbf{s}_i,
y_i^{\mathrm{out}}
\right)
\right\}_{i=1}^{N_{\mathrm{out}}}.
$
For clarity, we omit the trajectory index $i$ when discussing a single trajectory. Our goal is to learn a PRM that estimates the state validity of each reasoning prefix $\mathbf{s}_{\leq t}$ while jointly leveraging process and outcome supervision.

\paragraph{Independent Prefix Classification}
A common discriminative PRM formulation
\citep{lightman2024let,zhang2025lessons}
treats each reasoning prefix as an independent binary classification target.
For the $i$-th trajectory, a PRM with parameters $\theta$ predicts
$
q_{i,t}
=
p_{\theta}
\left(
y_{i,t}^{\mathrm{step}}=1
\mid
x_i,\mathbf{s}_{i,\leq t}
\right),
$
where $q_{i,t}$ denotes the predicted validity of the reasoning prefix after
step $s_{i,t}$. The model is trained on the available process annotations using
the binary cross-entropy objective
\begin{equation}
\mathcal{L}_{\mathrm{BCE}}
=
-\frac{1}{\sum_{i=1}^{N_{\mathrm{step}}} |\mathcal{A}_i|}
\sum_{i=1}^{N_{\mathrm{step}}}
\sum_{t\in\mathcal{A}_i}
\left[
y_{i,t}^{\mathrm{step}}\log q_{i,t}
+
\left(1-y_{i,t}^{\mathrm{step}}\right)
\log\left(1-q_{i,t}\right)
\right].
\label{eq:bce}
\end{equation}
Although each prediction is conditioned on the full reasoning prefix, the
objective decomposes over individual prefixes and imposes no explicit relation between their predicted validity across successive steps. Moreover, this objective relies exclusively on process annotations and cannot naturally incorporate outcome-only trajectories.

\subsection{Reasoning State Propagation}

To explicitly relate reasoning states across successive steps, we represent the state of each reasoning prefix as
$
z_t \in \mathcal{Z}=\{\mathrm{G},\mathrm{B}\}.
$
Here, $\mathrm{G}$ denotes that the reasoning currently maintained after step $s_t$ is valid and contains no unresolved error, whereas $\mathrm{B}$ denotes that an error remains unresolved in the current reasoning state. For each reasoning step $s_t$, we parameterize two probabilities that describe its effect on the preceding reasoning state:
\begin{align}
\alpha_t
&\coloneqq
p_{\theta}
\left(
z_t=\mathrm{B}
\mid
z_{t-1}=\mathrm{G},
x,\mathbf{s}_{<t},s_t
\right), \\
\beta_t
&\coloneqq
p_{\theta}
\left(
z_t=\mathrm{G}
\mid
z_{t-1}=\mathrm{B},
x,\mathbf{s}_{<t},s_t
\right).
\end{align}
Here, $\alpha_t$ denotes the \emph{break} probability that step $s_t$ introduces an unresolved error into a previously valid reasoning state, and $\beta_t$ denotes the \emph{repair} probability that step $s_t$ resolves or supersedes an earlier error, returning the current reasoning state to a valid one.
The break and repair probabilities define a local state propagation matrix for step $s_t$:
\begin{equation}
\mathbf{M}_t
=
\begin{bmatrix}
1-\alpha_t & \alpha_t \\
\beta_t & 1-\beta_t
\end{bmatrix}.
\end{equation}
The rows correspond to the preceding state $z_{t-1}$, while the columns correspond to the updated state $z_t$. $\mathbf{M}_t$ describes a one-step propagation between adjacent reasoning states, conditioned on the preceding reasoning prefix and the newly incorporated step $s_t$.
Let
$
\mathbf{p}_t
=
\begin{bmatrix}
p_t^{\mathrm{G}} & p_t^{\mathrm{B}}
\end{bmatrix}
$
denote the probability distribution over reasoning states after step $s_t$.
Before any reasoning step, we initialize the state as
$
\mathbf{p}_0=[1,0],
$
and propagate it as
\begin{equation}
\mathbf{p}_t
=
\mathbf{p}_{t-1}\mathbf{M}_t
=
\mathbf{p}_0
\prod_{j=1}^{t}\mathbf{M}_j.
\end{equation}
We use $p_t^{\mathrm{G}}$ as the process score for the reasoning prefix
$\mathbf{s}_{\leq t}$. Unlike the independently predicted score $q_t$,
$p_t^{\mathrm{G}}$ depends on all preceding state updates through sequential
propagation. Here,
$\mathrm{G}\!\rightarrow\!\mathrm{G}$ preserves valid reasoning,
$\mathrm{G}\!\rightarrow\!\mathrm{B}$ captures error introduction,
$\mathrm{B}\!\rightarrow\!\mathrm{B}$ captures error persistence, and
$\mathrm{B}\!\rightarrow\!\mathrm{G}$ captures error recovery.
The resulting state sequence therefore explicitly links reasoning states across successive steps.


\paragraph{Transition Parameterization}
To estimate the break and repair probabilities, we insert two special tokens, $\langle\mathrm{BREAK}\rangle$ and $\langle\mathrm{REPAIR}\rangle$, at each reasoning boundary. Boundary $0$ is placed after the question $x$, and boundary $t$ is placed after reasoning step $s_t$. Specifically, at each boundary $t$, we insert a pair of update tokens
$\mathbf{u}_t=
(\langle\mathrm{BREAK}\rangle_t,
\langle\mathrm{REPAIR}\rangle_t)$,
yielding
\[
\widetilde{\mathbf{s}}
=
(x,\mathbf{u}_0,s_1,\mathbf{u}_1,\ldots,s_T,\mathbf{u}_T).
\]
Our PRM is built on a pretrained LRM backbone $f_{\phi}$, which encodes $\widetilde{\mathbf{s}}$ in a single forward pass. Here, $\phi$ denotes the backbone parameters, including the embeddings of the newly introduced special tokens. Let
$\mathbf{h}_t^{\mathrm{br}}$ and $\mathbf{h}_t^{\mathrm{rp}}$
denote the hidden representations of
$\langle\mathrm{BREAK}\rangle_t$ and
$\langle\mathrm{REPAIR}\rangle_t$ at boundary $t$.
Since boundary $t$ follows step $s_t$, these representations encode the reasoning prefix $(x,\mathbf{s}_{\leq t})$. To characterize the state update induced by step $s_t$, we use the representations at the boundaries before and after $s_t$. The break and repair probabilities are computed as
\begin{align}
\alpha_t
&=
\sigma\left(
g_{\mathrm{br}}
\left(
\mathbf{h}_{t-1}^{\mathrm{br}}
\mathbin{\Vert}
\mathbf{h}_t^{\mathrm{br}}
\right)
\right), \\
\beta_t
&=
\sigma\left(
g_{\mathrm{rp}}
\left(
\mathbf{h}_{t-1}^{\mathrm{rp}}
\mathbin{\Vert}
\mathbf{h}_t^{\mathrm{rp}}
\right)
\right),
\end{align}
where $\mathbin{\Vert}$ denotes the vector concatenation.
$g_{\mathrm{br}}(\cdot)$ and
$g_{\mathrm{rp}}(\cdot)$ are two independent MLP heads, and $\sigma(\cdot)$ denotes the sigmoid function. All break and repair probabilities along the trajectory are obtained from a single forward pass. 
We denote all trainable parameters of \algo\ by $\theta$.


\subsection{Learning from Process and Outcome Supervision}

The propagated reasoning states enable \algo{} to naturally incorporate both process and outcome supervision. We define the corresponding learning objectives below.

For each process-annotated trajectory in $\mathcal{D}_{\mathrm{step}}$, we
supervise the propagated state at every annotated step
$t\in\mathcal{A}_i$. Since $y_{i,t}^{\mathrm{step}}=1$ indicates a valid reasoning step, we define
\begin{equation}
\mathcal{L}_{\mathrm{step}}
=
-\frac{1}{\sum_{i=1}^{N_{\mathrm{step}}}|\mathcal{A}_i|}
\sum_{i=1}^{N_{\mathrm{step}}}
\sum_{t\in\mathcal{A}_i}
\left[
y_{i,t}^{\mathrm{step}}
\log p_{i,t}^{\mathrm{G}}
+
\left(1-y_{i,t}^{\mathrm{step}}\right)
\log p_{i,t}^{\mathrm{B}}
\right].
\end{equation}
Unlike BCE in \eqref{eq:bce} applied to independently predicted scores $q_{i,t}$, this loss supervises state probabilities that depend on the preceding reasoning states through sequential propagation.


For each outcome-annotated trajectory in $\mathcal{D}_{\mathrm{out}}$, we define
\begin{equation}
\mathcal{L}_{\mathrm{out}}
=
-\frac{1}{N_{\mathrm{out}}}
\sum_{i=1}^{N_{\mathrm{out}}}
\left[
y_i^{\mathrm{out}}
\log p_{i,T_i}^{\mathrm{G}}
+
\left(1-y_i^{\mathrm{out}}\right)
\log p_{i,T_i}^{\mathrm{B}}
\right].
\end{equation}
Because the final state $\mathbf{p}_{i,T_i}$ depends on all state updates along the trajectory, the outcome loss is backpropagated through the propagation process and
therefore trains the break and repair predictions at intermediate steps.
We jointly optimize the two supervision signals as
$
\mathcal{L}_{\mathrm{total}}
=
\mathcal{L}_{\mathrm{step}}
+
\mathcal{L}_{\mathrm{out}}.
$





\paragraph{Effect of Outcome Supervision on State Transitions}
Process labels directly supervise intermediate states, whereas outcome supervision reaches earlier transitions through state propagation. For an outcome-annotated trajectory, let $c\in\{\mathrm{G},\mathrm{B}\}$ denote the terminal target and define
$h_t(a)=p_\theta(z_T=c\mid z_t=a,x,\mathbf{s})$ as the likelihood of reaching $c$ from state $a$ through the remaining reasoning. Let
$Z_{\mathrm{out}}=p_T^c$ and $\ell^{\mathrm{out}}=-\log Z_{\mathrm{out}}$. Then, with respect to the pre-sigmoid logits of $\alpha_t$ and $\beta_t$,
\[
\frac{\partial \ell^{\mathrm{out}}}
{\partial \operatorname{logit}(\alpha_t)}
=
\frac{p_{t-1}^{\mathrm{G}}\alpha_t(1-\alpha_t)}
     {Z_{\mathrm{out}}}
\bigl(h_t(\mathrm{G})-h_t(\mathrm{B})\bigr),
\quad
\frac{\partial \ell^{\mathrm{out}}}
{\partial \operatorname{logit}(\beta_t)}
=
\frac{p_{t-1}^{\mathrm{B}}\beta_t(1-\beta_t)}
      {Z_{\mathrm{out}}}
\bigl(h_t(\mathrm{B})-h_t(\mathrm{G})\bigr).
\]

These gradients show how the final outcome shapes intermediate state transitions. If $h_t(\mathrm{G})>h_t(\mathrm{B})$, the outcome loss favors transitions toward $\mathrm{G}$ by decreasing \emph{break} and increasing \emph{repair}, with the opposite effect when $h_t(\mathrm{B})>h_t(\mathrm{G})$. Detailed analysis is provided in Appendix~\ref{app:rsp_supervision_analysis}.

\section{Experiments}
\label{exps}

\subsection{Experimental Settings}

\paragraph{Model Architecture}
We build \algo{} on the Qwen3 family, using Qwen3-4B as the default backbone unless otherwise specified. Following the formulation of \algo{}, we use two independent MLP heads to predict the \emph{break} and \emph{repair} probabilities. The backbone parameters, the embeddings of the $\langle\mathrm{BREAK}\rangle$ and $\langle\mathrm{REPAIR}\rangle$ tokens, and the two MLP heads are optimized during training.

\paragraph{Training Data}
We train \algo{} using both process and outcome supervision. For process supervision, we use PRM800K~\citep{lightman2024let}, which contains human process annotations for trajectories. Unless otherwise specified, we randomly select 20\% of PRM800K as the process-annotated training set. For outcome supervision, we use the filtered correctness labels from AceMath-RM~\citep{liu2025acemathadvancingfrontiermath} as outcome supervision for the terminal reasoning state. AceMath-RM ranks and filters candidate responses with an LLM judge to retain high-confidence examples, reducing cases where final answer correctness is inconsistent with the validity of the terminal reasoning state. Each batch contains process-annotated and outcome-annotated examples at a default ratio of $1{:}3$. We train models for two epochs, where one epoch corresponds to a complete pass over the process-annotated set, and outcome-annotated examples are sampled to maintain the ratio.

\paragraph{Baselines}
We compare \algo{} with representative process- and outcome-supervised reward models. Supervised PRM~\citep{lightman2024let} independently predicts the validity of each reasoning prefix using process annotations. OVM~\citep{yu2024ovm} uses outcome supervision by assigning the outcome as the target for each reasoning step. We additionally compare with Qwen2.5-Math-PRM~\citep{zhang2025lessons}, using its 7B checkpoint by default, and CRM~\citep{zhang2025linking} as another representative PRM. To compare different ways of combining process and outcome supervision, we construct additional baselines: Pseudo-Label PRM, Joint-Supervised PRM, and CRM$^{\dagger}$. Pseudo-Label PRM follows a standard semi-supervised strategy \citep{sohn2020fixmatch}, using high-confidence predictions as pseudo process labels. Joint-Supervised PRM extends Supervised PRM by additionally supervising the prediction at the final reasoning prefix with the outcome label using BCE loss. CRM$^{\dagger}$ extends CRM with an additional outcome objective, allowing it to incorporate both process and outcome supervision. More details are provided in Appendix~\ref{app:experimental_details}.


\subsection{Process-Guided Beam Search}
\label{exp:beam_search}

We evaluate whether process rewards can reliably rank intermediate reasoning prefixes in beam search. We consider PRMs built on Qwen3-1.7B and Qwen3-4B and evaluate them on MATH500~\citep{lightman2024let} and Gaokao~\citep{liao2024mario}. The policy model is fixed to Qwen3-1.7B for all experiments. We fix the beam width to $b=4$ and vary the expansion width $e\in\{4,8,12\}$. At each search step, each of the $b$ retained prefixes is expanded into $e$ candidate continuations. The resulting $b\times e$ candidates are ranked by the PRM, and the top-$b$ prefixes are retained for the next step.

\begin{table*}[t]
\centering
\caption{
Beam search accuracy (\%) on MATH500 and Gaokao. The beam width is fixed to $4$, and $e$ denotes the expansion width. The best and second-best results are shown in \textbf{bold} and \underline{underlined}.}
\label{tab:beam_search}
\renewcommand{\arraystretch}{1.1}
\resizebox{\textwidth}{!}{

\begin{tabular}{llcccccc}
\toprule
\multirow{2}{*}{PRM Backbone}
&
\multirow{2}{*}{Reward Model}
&
\multicolumn{3}{c}{MATH500}
&
\multicolumn{3}{c}{Gaokao}
\\
\cmidrule(lr){3-5}
\cmidrule(lr){6-8}
&
&
$e=4$
&
$e=8$
&
$e=12$
&
$e=4$
&
$e=8$
&
$e=12$
\\
\midrule

Qwen2.5-Math-7B
& Qwen2.5-Math-PRM
& 74.8 & 74.0 & 75.8 & 58.5 & 55.6 & 59.3 \\
\midrule

\multirow{5}{*}{Qwen3-1.7B}
& Supervised PRM
& 54.8 & 53.6 & 55.4 & 44.0 & 45.7 & 48.3 \\
& OVM
& \underline{67.9} & \underline{68.4} & \underline{69.5} & 54.5 & \underline{58.3} & \underline{58.3} \\
& Pseudo-Label PRM
& 65.3 & 64.8 & 67.7 & 53.1 & 52.5 & 55.0 \\
& Joint-Supervised PRM
& 67.4 & 64.6 & 64.0 & 54.4 & 57.2 & 55.9 \\
& CRM$^{\dagger}$
& 62.2 & 66.2 & 63.4 & \underline{55.6} & 53.5 & 56.6 \\
& \textbf{\algo{}}
& \textbf{68.8} & \textbf{70.5} & \textbf{70.0} & \textbf{55.6} & \textbf{58.5} & \textbf{59.3} \\
\midrule

\multirow{5}{*}{Qwen3-4B}
& Supervised PRM
& 62.3 & 59.8 & 61.4 & 52.6 & 53.3 & 52.7 \\
& OVM
& 73.1 & 75.2 & 74.9 & 61.7 & \underline{64.8} & \underline{63.0} \\
& Pseudo-Label PRM
& 68.2 & 64.4 & 65.7 & 56.4 & 49.9 & 53.4 \\
& Joint-Supervised PRM
& \underline{75.4} & \underline{75.4} & \underline{75.6} & \underline{62.9} & 63.2 & 61.1 \\
& CRM$^{\dagger}$
& 68.5 & 68.2 & 69.9 & 55.3 & 55.3 & 56.8 \\
& \textbf{\algo{}}
& \textbf{76.0} & \textbf{78.6} & \textbf{79.8} & \textbf{63.5} & \textbf{65.8} & \textbf{67.9} \\
\bottomrule
\end{tabular}
}
\end{table*}

\paragraph{Analysis}
Table~\ref{tab:beam_search} shows that \algo{} achieves the strongest overall beam search performance across both Qwen3 PRM backbones and expansion widths. The advantage is particularly pronounced with the Qwen3-4B backbone at larger expansion widths. At $e=12$, \algo{} outperforms OVM by 4.9\% on both MATH500 and Gaokao. OVM remains competitive, suggesting that outcome supervision alone can provide useful signals for prefix ranking. In contrast, Pseudo-Label PRM performs substantially worse and exhibits larger variation across expansion widths, particularly on Gaokao. Consistent gains over Joint-Supervised PRM and CRM$^{\dagger}$ further show the benefit of reasoning state propagation when jointly leveraging process and outcome supervision.

\subsection{Best-of-$N$ Selection}
\label{exp:bon}
Best-of-$N$ evaluates whether a reward model can select a correct response from multiple sampled candidates. For each generator and question, we sample $N$ responses and select the response with the highest final score. We consider $N\in\{8,16,32,64,128\}$ and report the average accuracy in these settings. All evaluations are conducted on MATH500. For PRMs, we use the process score of the final reasoning step as the response score. We consider six generators from the LLaMA, Qwen, and DeepSeek model families. 

\paragraph{Analysis}
Table~\ref{tab:bon} shows that \algo{} achieves the highest average
Best-of-$N$ accuracy across the six generators. In particular, \algo{}
slightly exceeds Qwen2.5-Math-PRM in average accuracy despite the latter using additional large-scale process supervision. Adding outcome supervision to CRM improves its average accuracy from 72.1\% to 73.4\%, while \algo{} further improves it to 74.4\%. In contrast, Pseudo-Label PRM performs substantially worse, indicating that pseudo process labels provide a less effective way to combine the two supervision sources. Joint-Supervised PRM is also competitive but remains below \algo{}, showing that it does not fully exploit the two supervision signals. The results remain competitive across LLaMA, Qwen, and DeepSeek generators, indicating that the benefit of \algo{} is not limited to a single generator family.

\begin{table*}[t]
\centering
\caption{Average Best-of-$N$ accuracy (\%) over $N\in\{8,16,32,64,128\}$. Proc./Out. denote process/outcome supervision, and $\checkmark^{+}$ indicates additional large-scale process supervision.}
\label{tab:bon}

\renewcommand{\arraystretch}{1.1}

\resizebox{\textwidth}{!}{
\begin{tabular}{lccccccccc}
\toprule
Method
& Proc.
& Out.
& \shortstack{LLaMA3\\70B}
& \shortstack{LLaMA3.3\\70B}
& \shortstack{Qwen3\\1.7B}
& \shortstack{Qwen3\\4B}
& \shortstack{Qwen3\\8B}
& \shortstack{DeepSeek-R1-0528\\Qwen3-8B}
& Avg.
\\
\midrule

Pass@1
& -- & --
& 43.6 & 69.0 & 65.8 & 77.4 & 79.2 & 51.6 & 64.4 \\

Supervised PRM
& $\checkmark$ & --
& 57.2 & 74.6 & 76.0 & 80.8 & 79.9 & 46.0 & 69.1 \\

OVM
& -- & $\checkmark$
& 60.5 & 75.5 & 77.9 & \underline{84.1} & \underline{83.2} & \underline{62.4} & 73.9 \\

CRM
& $\checkmark$ & --
& 58.2 & \textbf{77.8} & \underline{79.1} & 83.1 & 82.5 & 52.1 & 72.1 \\

Qwen2.5-Math-PRM
& $\checkmark^{+}$ & --
& 60.6 & 77.1 & \textbf{79.9} & 83.5 & 82.4 & 61.0 & \underline{74.1} \\

Pseudo-Label PRM
& $\checkmark$ & $\checkmark$
& 43.6 & 69.2 & 65.8 & 77.6 & 80.2 & 59.9 & 66.1 \\

Joint-Supervised PRM
& $\checkmark$ & $\checkmark$
& \underline{60.7} & 77.1 & 78.8 & 82.4 & 82.9 & 59.9 & 73.6 \\

CRM$^{\dagger}$
& $\checkmark$ & $\checkmark$
& 60.6 & \underline{77.4} & 78.6 & 82.0 & 81.9 & 60.0 & 73.4 \\

\textbf{\algo{}}
& $\checkmark$ & $\checkmark$
& \textbf{61.2} & 76.1 & 78.3 & \textbf{84.1}
& \textbf{83.3} & \textbf{63.3} & \textbf{74.4} \\

\bottomrule
\end{tabular}
}
\end{table*}


\subsection{Reinforcement Learning with Process Rewards}
\label{exp:rl}

We further evaluate whether process rewards can provide effective fine-grained feedback for policy optimization. Following the verifier-gated strategy of VeriGate~\citep{agrawal2026verigate}, we use PRM feedback only when all sampled responses for a prompt receive zero verifier reward, while retaining standard verifier-based GRPO updates otherwise. Unlike VeriGate, we directly use the PRM score at each reasoning step as the step-level reward without aggregating rewards from subsequent steps. We train on 8K prompts from DAPO-MATH-17K~\citep{yu2026dapo} and evaluate the resulting policy using zero-shot Avg@16 accuracy on six mathematical reasoning benchmarks.


\begin{table*}[t]
\centering
\caption{
Avg@16 accuracy (\%) after reinforcement learning.
All methods use the same policy initialization, training data, and optimization budget.}
\label{tab:rl_results}

\renewcommand{\arraystretch}{1.12}

\resizebox{\textwidth}{!}{
\begin{tabular}{lcccccc@{\hspace{10pt}}c}
\toprule
\textbf{Method}
& \textbf{AIME}
& \textbf{AMC}
& \textbf{GSM8K}
& \textbf{MATH500}
& \textbf{Minerva}
& \textbf{Olympiad}
& \textbf{Avg.} \\
\midrule

Base Policy
& 23.3 & 70.0 & 92.5 & 78.6 & 38.2 & 52.4 & 59.2 \\

GRPO
& 23.3 & 72.3 & \textbf{94.2} & \underline{88.2} & \textbf{40.2} & 59.6 & 63.0 \\

\midrule

Supervised PRM
& 16.7 & 68.3 & 92.3 & 83.4 & 38.2 & 54.8 & 59.0 \\

OVM
& 31.1 & 71.3 & 92.6 & 86.2 & 37.6 & 58.4 & 62.9 \\

Qwen2.5-Math-PRM
& 31.1 & 68.3 & \underline{93.8} & 86.8 & \underline{39.8} & 59.8 & \underline{63.3} \\

Pseudo-Label PRM
& 26.7 & \underline{72.3} & 92.6 & 85.8 & 35.8 & 57.3 & 61.8 \\

Joint-Supervised PRM
& \underline{31.1} & 70.0 & 92.3 & 87.2 & 38.2 & \underline{59.8} & 63.1 \\

CRM$^{\dagger}$
& 23.3 & 70.0 & 92.5 & 86.2 & 37.0 & 57.3 & 61.1 \\

\midrule

\textbf{\algo{}}
& \textbf{38.9}
& \textbf{72.5}
& 92.9
& \textbf{88.4}
& 39.3
& \textbf{60.2}
& \textbf{65.4} \\

\bottomrule
\end{tabular}
}
\end{table*}

\paragraph{Analysis}
Table~\ref{tab:rl_results} shows that \algo{} achieves the strongest overall performance, improving the average accuracy by 2.1\% over PRM-guided with Qwen2.5-Math-PRM. The gain is particularly pronounced on AIME, where \algo{} outperforms Qwen2.5-Math-PRM by 7.8\%. Supervised PRM reduces the average accuracy below that of the base policy, while OVM, Pseudo-Label PRM, and CRM$^{\dagger}$ all remain below GRPO on average. \algo{} is the only PRM-guided method that substantially improves the average accuracy over GRPO, indicating that introducing process rewards alone does not necessarily benefit policy optimization and that the quality of the process signal is critical. These results highlight the benefit of process rewards from \algo{} for policy optimization.


\subsection{Analysis of Reasoning State Propagation}
\label{exp:rsp_analysis}
\begin{figure*}[!t]
    \centering

    \begin{minipage}[t]{0.33\textwidth}
        \centering
        \includegraphics[width=\linewidth]{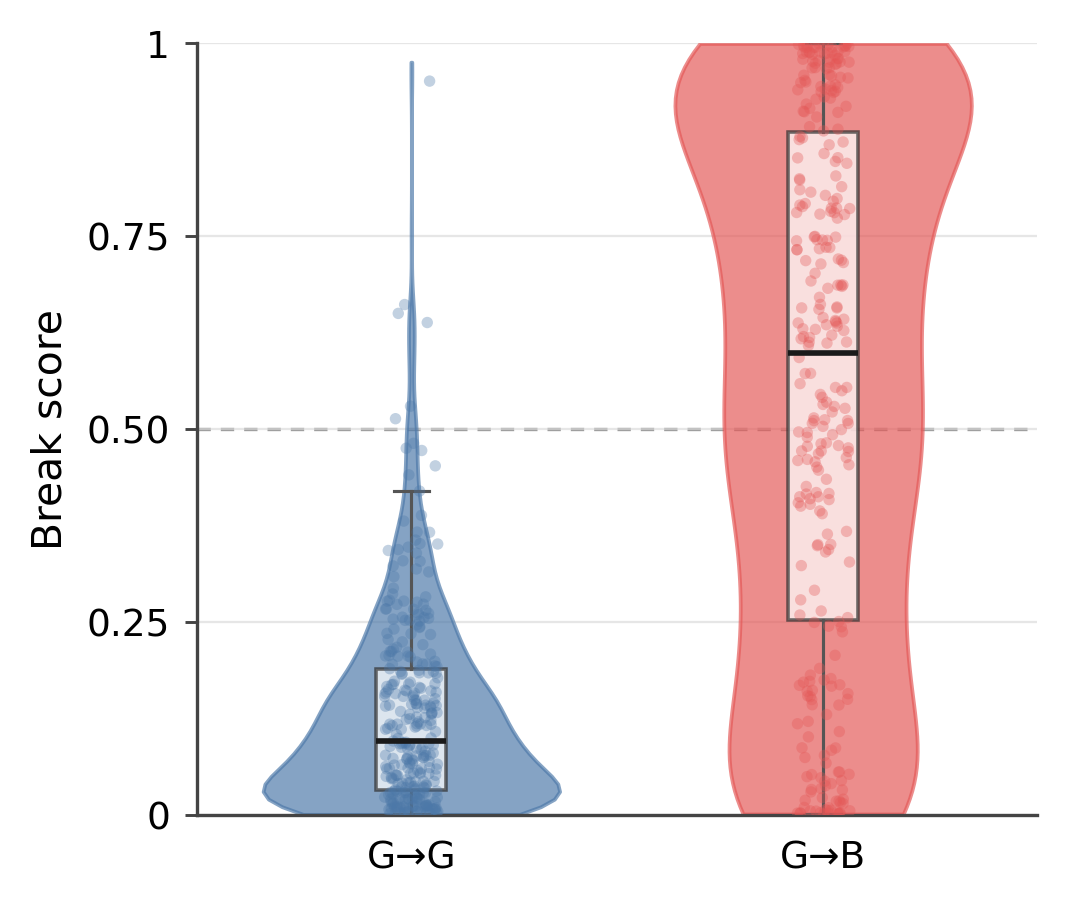}
        \vspace{-2mm}
        \centerline{\small (a) Break prediction}
    \end{minipage}
    \hfill
    \begin{minipage}[t]{0.33\textwidth}
        \centering
        \includegraphics[width=\linewidth]{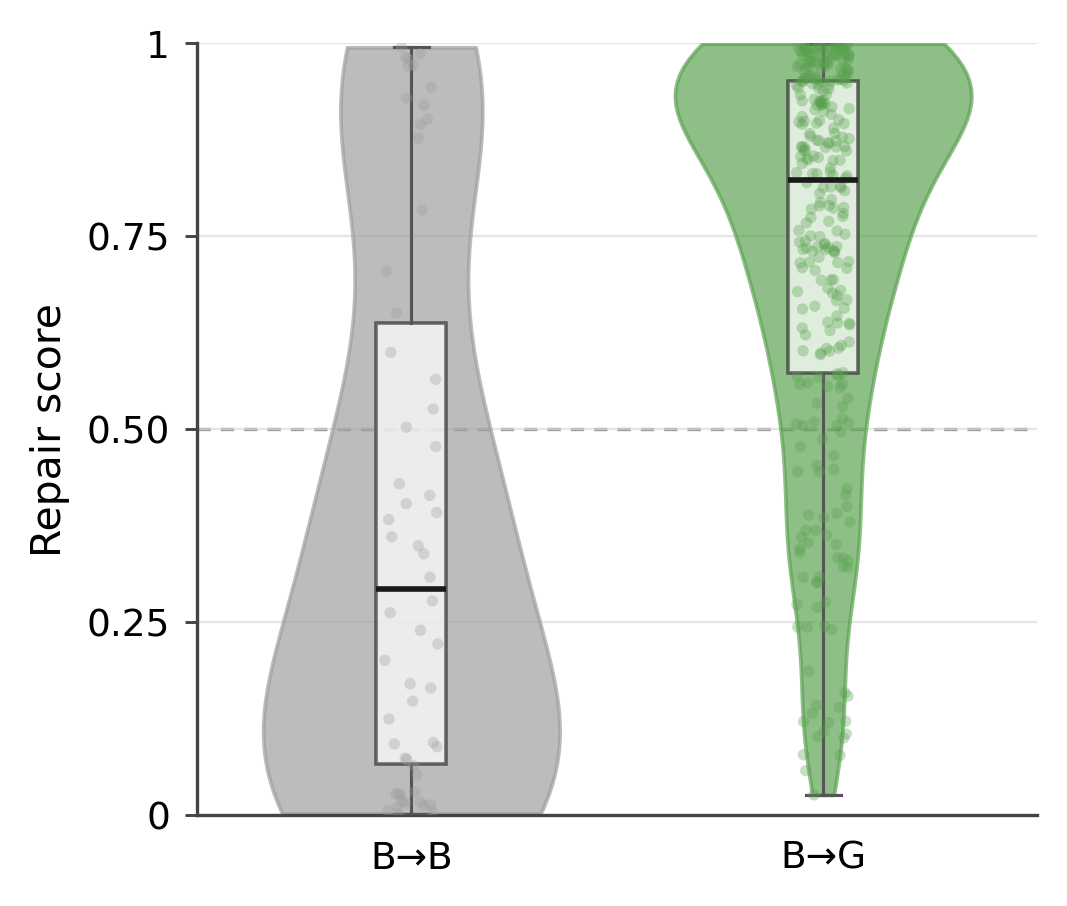}
        \vspace{-2mm}
        \centerline{\small (b) Repair prediction}
    \end{minipage}
    \hfill
    \begin{minipage}[t]{0.31\textwidth}
        \centering
        \includegraphics[width=\linewidth]{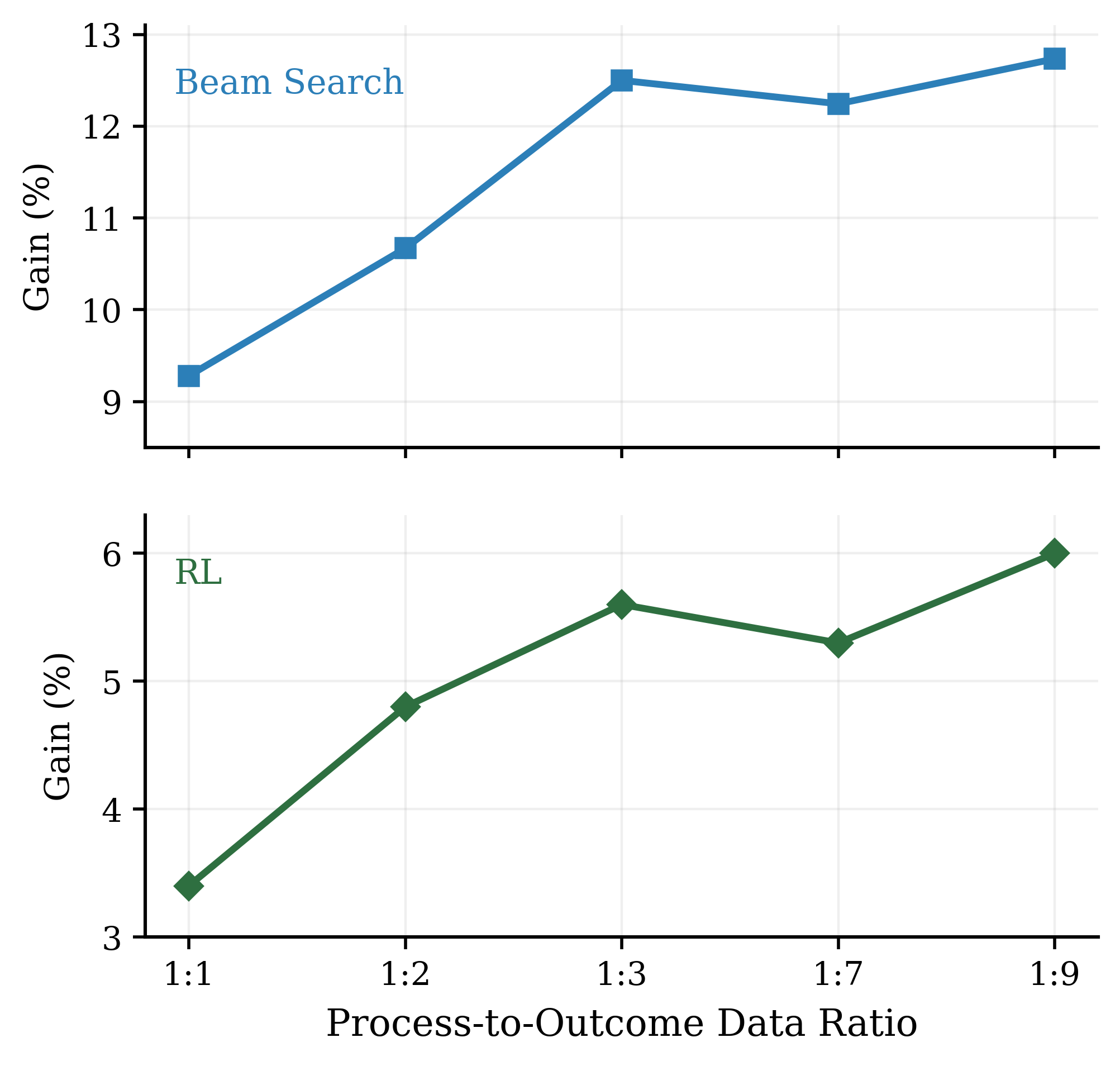}
        \vspace{-2mm}
        \centerline{\small (c) Outcome supervision scaling}
    \end{minipage}

    \caption{
    (a) Distributions of predicted \emph{break} scores for
    $\mathrm{G}\!\rightarrow\!\mathrm{G}$ and
    $\mathrm{G}\!\rightarrow\!\mathrm{B}$.
    (b) Distributions of predicted \emph{repair} scores for
    $\mathrm{B}\!\rightarrow\!\mathrm{B}$ and
    $\mathrm{B}\!\rightarrow\!\mathrm{G}$.
    (c) Performance gains over Process-Only under increasing amounts of outcome supervision for beam search and reinforcement learning.
    }
    \label{fig:rsp_analysis}
\end{figure*}

\paragraph{Transition Predictions}
We examine whether the predicted transitions reflect meaningful changes in reasoning states. In particular, we compare the score distributions under the four state update patterns
$\mathrm{G}\!\rightarrow\!\mathrm{G}$,
$\mathrm{G}\!\rightarrow\!\mathrm{B}$,
$\mathrm{B}\!\rightarrow\!\mathrm{B}$, and
$\mathrm{B}\!\rightarrow\!\mathrm{G}$. Publicly available process-annotated data contain relatively few examples after reasoning has entered an invalid state, making $\mathrm{B}\!\rightarrow\!\mathrm{B}$ and
$\mathrm{B}\!\rightarrow\!\mathrm{G}$ cases particularly scarce. We therefore use Qwen3-32B as an LLM judge to determine the validity of each reasoning step and construct approximately 10K step pairs covering the four state update patterns. As shown in Figure~\ref{fig:rsp_analysis}, for \emph{break} predictions, $\mathrm{G}\!\rightarrow\!\mathrm{G}$ cases are concentrated at low scores, whereas $\mathrm{G}\!\rightarrow\!\mathrm{B}$ cases receive substantially higher scores. Similarly, \emph{repair} scores are generally higher for $\mathrm{B}\!\rightarrow\!\mathrm{G}$ than for $\mathrm{B}\!\rightarrow\!\mathrm{B}$. The overall separation indicates that the learned transition scores capture meaningful changes in reasoning states.


\paragraph{Scaling with Outcome Supervision}
We further study how \algo{} benefits from increasing the amount of outcome supervision while keeping process supervision fixed. We vary the process-to-outcome data ratio from $1{:}1$ to $1{:}9$ and report the performance gain over the corresponding Process-Only model. Figure~\ref{fig:rsp_analysis}(c) shows that adding outcome supervision provides clear gains over Process-Only for both beam search and reinforcement learning. The gains generally increase as more outcome-annotated samples are introduced, showing that \algo{} can effectively benefit from additional outcome supervision for reasoning search and policy optimization.



\subsection{Ablation Study}
\label{exp:ablation}

Table~\ref{tab:ablation} examines the contributions of the two supervision and the main design choices of \algo{}. Except for the results reported in Table~\ref{tab:beam_search}, all subsequent beam search results use the Qwen3-4B PRM backbone with $b=4$ and $e=8$, and report the average accuracy on MATH500 and Gaokao.


\paragraph{Supervision Sources}
\emph{Process-Only} and \emph{Outcome-Only} use only process and outcome supervision, respectively. Process-Only shows large drops on beam search and ProcessBench~\citep{zheng2024processbench}, whereas Outcome-Only nearly preserves Best-of-$N$ performance but drops by 9.6\% on ProcessBench. This contrast suggests that outcome supervision provides strong signals for response selection and search, while process annotations remain important for fine-grained process evaluation. Combining the two sources of supervision gives the best performance in the evaluated settings.

\paragraph{Repair Modeling}
\emph{No Repair} sets the repair probability to zero during training and inference, so an invalid reasoning state cannot return to a valid state. Removing repair degrades performance across all four settings, with the largest drop of 3.3\% on beam search. This result shows that explicitly modeling error recovery is important for reasoning state propagation.

\begin{table}[!t]
\centering
\caption{
Ablation study across evaluation settings.
Colored numbers indicate changes relative to \textbf{Ours},
with darker colors denoting larger degradation.
}
\label{tab:ablation}

\renewcommand{\arraystretch}{1.1}

\begin{tabular}{lcccc}
\toprule
\textbf{Variant}
& \textbf{BoN}
& \makecell{\textbf{Beam} \textbf{Search}}
& \makecell{\textbf{Process}\textbf{Bench}}
& \textbf{RL} \\
\midrule

Process-Only
& \dcell{72.0}{-2.4}{dropMedium}
& \dcell{59.7}{-12.5}{dropSevere}
& \dcell{58.3}{-7.5}{dropSevere}
& \dcell{59.8}{-5.6}{dropLarge}
\\

Outcome-Only
& \dcell{74.2}{-0.2}{dropSmall}
& \dcell{69.5}{-2.7}{dropMedium}
& \dcell{56.2}{-9.6}{dropSevere}
& \dcell{61.6}{-3.8}{dropMedium}
\\

\midrule

No Repair
& \dcell{73.6}{-0.8}{dropSmall}
& \dcell{68.9}{-3.3}{dropLarge}
& \dcell{64.5}{-1.3}{dropMedium}
& \dcell{63.2}{-2.2}{dropMedium}
\\

Current Representation Only
& \dcell{73.5}{-0.9}{dropSmall}
& \dcell{70.1}{-2.1}{dropMedium}
& \dcell{64.2}{-1.6}{dropMedium}
& \dcell{62.3}{-3.1}{dropMedium}
\\

Shared Token
& \dcell{73.5}{-0.9}{dropSmall}
& \dcell{69.7}{-2.5}{dropMedium}
& \dcell{63.5}{-2.3}{dropMedium}
& \dcell{63.8}{-1.6}{dropMedium}
\\

No Outcome Propagation
& \dcell{73.9}{-0.5}{dropSmall}
& \dcell{65.5}{-6.7}{dropSevere}
& \dcell{61.8}{-4.0}{dropLarge}
& \dcell{64.2}{-1.2}{dropMedium}
\\

\midrule

\rowcolor{oursBG}
\textbf{\algo{}}
& \textbf{74.4}
& \textbf{72.2}
& \textbf{65.8}
& \textbf{65.4}
\\

\bottomrule
\end{tabular}
\end{table}

\paragraph{Adjacent Representations}
\emph{Current Representation Only} predicts the break and repair probabilities using only the representation at the current reasoning boundary, rather than the representations before and after the reasoning step. The consistent degradation shows that using both boundary representations provides more informative signal for estimating how the step updates reasoning states.

\paragraph{Separate Break and Repair Tokens}
\emph{Shared Token} replaces the separate
$\langle\mathrm{BREAK}\rangle$ and $\langle\mathrm{REPAIR}\rangle$ tokens with a single shared token. The resulting performance drops across all four settings, particularly 2.5\% drop on beam search and 2.3\% on ProcessBench, indicating that separate token representations help the model distinguish the roles of break and repair, while sharing a single token may introduce semantic interference between two conceptually different state transitions.

\paragraph{Outcome Propagation}
We further examine whether outcome supervision benefits intermediate state learning through the propagation mechanism. Specifically, we preserve the same forward state propagation while blocking backpropagation from the terminal outcome loss to earlier state transitions. This variant consistently degrades performance, with particularly large drops of 6.7\% on beam search and 4.0\% on ProcessBench. Since the forward propagation remains unchanged, the results show that propagating outcome supervision to earlier state updates is important for learning effective process rewards.

\section{Conclusion}
\label{conclusion}

We introduced \algo{}, a reasoning state propagation approach for process reward modeling. By modeling \emph{break} and \emph{repair} probabilities, \algo{} explicitly links reasoning states across successive steps and captures error introduction, persistence, and recovery. The propagated reasoning state also provides a natural way to combine scarce process supervision with scalable outcome supervision, allowing terminal outcome supervision to provide learning signals to earlier state updates. Our results demonstrate the effectiveness of this formulation and highlight the complementary roles of process and outcome supervision in learning
reliable process rewards.

\subsection*{AI use statement}
We used generative AI tools primarily to assist with language editing and improving the clarity and presentation of the manuscript. Generative AI was also used in a limited experimental analysis, where Qwen3-32B served as an LLM judge as described in Section~\ref{exp:rsp_analysis}. All AI-assisted content and outputs were reviewed and verified by the authors, who take full responsibility for the final content of the paper.

\subsection*{Reproducibility statement}
We provide the main implementation and experimental details in
Section~\ref{exps} and Appendix~\ref{app:experimental_details}, including training data, hyperparameters, evaluation settings, and baseline implementations. Code is provided in the supplementary material.





\bibliography{iclr2027_conference}
\bibliographystyle{iclr2027_conference}

\appendix

\section{Additional Experimental Details}
\label{app:experimental_details}

\subsection{\algo{} Training Details}
\label{app:reward_model_training}

Unless otherwise specified, we initialize \algo{} from Qwen3-4B and jointly optimize the backbone, the embeddings of the two special tokens, and the break and repair MLP heads. Following Section~\ref{method}, each head receives the concatenation of the boundary representations before and after a reasoning step. We randomly select 20\% of PRM800K and mix its process-annotated trajectories with outcome-annotated trajectories from AceMath-RM at a ratio of $1{:}3$. Training proceeds for two epochs over the selected PRM800K subset using AdamW with a learning rate of $5\times10^{-6}$. We optimize only the process and outcome objectives in Section~\ref{method}. Training is conducted in bfloat16 across eight NVIDIA H100 GPUs. We use the final checkpoint for evaluation. The code is provided in supplementary material.

\paragraph{Transition Heads.}
The break and repair predictors use two independent MLP heads with the same architecture. Given backbone hidden size $d$, each head takes the concatenated adjacent boundary representations in $\mathbb{R}^{2d}$ and applies a two-layer MLP, $2d \rightarrow d \rightarrow 1$, with GELU activation and dropout of $0.1$ after the input and hidden layer. The resulting scalar logits are converted to break and repair probabilities using a sigmoid.

\subsection{Reinforcement Learning Details}
\label{app:rl_details}

We initialize the policy from Qwen3-4B and perform full parameter training for one epoch on the 8K subset of DAPO-MATH-17K. For each prompt, we sample eight responses with a temperature of $1.0$ and top-$p$ of $1.0$. We use a rollout batch size of 16 and a training batch size of 128, with corresponding micro-batch sizes of 4 and 2. The maximum prompt and generation lengths are 512 and 4096 tokens, respectively. The actor learning rate is $1\times10^{-6}$ with a cosine schedule, and training is performed in bfloat16. During training, if all responses receive zero reward, we use \algo{} to score reasoning steps in each response. The step scores are normalized within the sampled group and assigned to the tokens of their corresponding steps, without accumulating scores from subsequent steps. Otherwise, the update uses only the outcome rewards. We use the Dr.~GRPO \citep{liu2025understanding} advantage estimator with a zero KL coefficient. All RL methods use the same policy training settings.

\subsection{Implementation Details of Additional Baselines}
\label{app:additional_baseline_details}

For a fair comparison, Pseudo-Label PRM, Joint-Supervised PRM, and CRM$^{\dagger}$ use the same backbone, training data mixture, and optimization settings as described above. Each baseline uses a single scalar step head evaluated at a \texttt{[PRM]} token appended after each reasoning step.

\paragraph{Pseudo-Label PRM}
For process-annotated trajectories, we apply BCE loss to all steps with available binary process labels. For outcome-annotated trajectories, the intermediate steps are treated as unlabeled, and hard pseudo labels are generated online from the current step predictions. A step is assigned a positive pseudo label when its predicted probability is at least $0.9$ and a negative pseudo label when the probability is at most $0.1$. The predictions between the two thresholds are ignored. Pseudo-labeling is enabled from the beginning of training.

\paragraph{Joint-Supervised PRM}
We use the same BCE loss as Supervised PRM on process-annotated trajectories. For each outcome-annotated trajectory, we additionally align the prediction at its final \texttt{[PRM]} position with the binary outcome label using BCE loss. Predictions at earlier steps of these trajectories are not supervised, and no pseudo labels are generated.

\paragraph{CRM$^{\dagger}$}
Following CRM~\citep{zhang2025linking}, the scalar prediction at step $t$ parameterizes the conditional probability $h_t$ that the trajectory first enters an invalid state at that step. The probability of remaining valid through step $T$ is
\begin{equation}
S_T=\prod_{t=1}^{T}(1-h_t).
\end{equation}
For a process-annotated trajectory, we set its trajectory label to positive when all available step labels are valid and to negative otherwise. For a negative trajectory, the first step labeled invalid is used as the observed first-error position $z$, and CRM optimizes both the trajectory outcome likelihood and the localization likelihood
$\Pr(z)=h_z\prod_{t<z}(1-h_t)$.
CRM$^{\dagger}$ further applies the outcome objective to outcome-annotated trajectories without process labels: it minimizes $-\log S_T$ for a positive outcome and $-\log(1-S_T)$ for a negative outcome, thereby marginalizing over the unknown first-error position.

\subsection{Outcome Data Construction}
\label{app:outcome_data_construction}

We convert the original AceMath-RM~\citep{liu2025acemathadvancingfrontiermath} reward model training data into the outcome-annotated trajectories used in our experiments. For each example, we extract the user question and the corresponding assistant response, and retain the original binary correctness label as the trajectory-level outcome. We remove multiple-choice examples, malformed conversations, and responses for which reliable reasoning steps or a final answer cannot be identified.

\paragraph{Outcome label reliability.}
The outcome labels in AceMath-RM are constructed with multiple stages of quality control. Candidate responses are first labeled as correct or incorrect by comparing their answers against reference labels using the Qwen-Math evaluation toolkit. AceMath-RM further ranks positive and negative candidates using Qwen2.5-Math-RM-72B and applies score-sorted sampling to retain high-confidence responses, reducing label noise introduced by heuristic answer matching \citep{liu2025acemathadvancingfrontiermath}. This ranking and filtering also helps reduce cases where a response reaches the correct final answer despite containing unresolved reasoning errors. We do not assume that a correct final answer implies that all preceding reasoning steps are valid. Instead, the outcome label is used only to supervise the final propagated state, while intermediate states are determined through state propagation and process supervision. Therefore, the filtered outcome labels provide reliable supervision for the final state in most cases.



\section{Process Error Identification}
\label{exp:processbench}

We evaluate process error identification on
ProcessBench~\citep{zheng2024processbench}. ProcessBench requires a model to identify the first erroneous reasoning step or determine that the entire solution is correct. For \algo{}, we consider a reasoning prefix valid when $p_t^{\mathrm{G}}\geq0.5$ and predict the first step with $p_t^{\mathrm{G}}<0.5$ as the error location. If no reasoning step is identified as erroneous, the solution is predicted to be fully correct. Following ProcessBench, we report F1 as the harmonic mean of the accuracy on erroneous solutions and the accuracy on fully correct solutions.

\paragraph{Analysis}
Table~\ref{tab:processbench} shows that, among the models trained under our shared experimental setup, \algo{} improves Supervised PRM by 9.2\%. \algo{} also outperforms Qwen2.5-Math-PRM on MATH by 4.2\%, although Qwen2.5-Math-PRM achieves the strongest overall result with additional large-scale process supervision. These results indicate that process annotations remain particularly important for ProcessBench, which explicitly evaluates fine-grained process error identification. The very low F1 of Pseudo-Label PRM further shows the difficulty of obtaining reliable fine-grained process supervision from pseudo labels.


\begin{table}[t]
\centering
\caption{
F1 scores (\%) on ProcessBench.
Our method classifies a step as valid when $p_t^{\mathrm{G}}\geq0.5$.
The last column reports the macro-average across all subsets.
}
\label{tab:processbench}
\setlength{\tabcolsep}{4.0pt}
\renewcommand{\arraystretch}{1.1}
\begin{tabular}{lccccc}
\toprule
Method
& GSM8K
& MATH
& OlympiadBench
& Omni-MATH
& Avg. \\
\midrule
Supervised PRM
& 69.8 & 60.3 & 48.5 & 47.9 & 56.6 \\
OVM
& 49.9 & 44.1 & 28.1 & 28.2 & 37.6 \\
Qwen2.5-Math-PRM
& \textbf{81.8} & 67.3 & \textbf{66.8} & \textbf{66.2} & \textbf{70.5} \\
Pseudo-Label PRM
& 1.9 & 2.1 & 2.0 & 2.2 & 2.1 \\
Joint-Supervised PRM
& 70.1 & 66.2 & 57.0 & 52.8 & 61.5 \\
CRM$^{\dagger}$
& \underline{73.8} & 56.2 & 30.7 & 24.4 & 46.3 \\
\midrule
\textbf{\algo{}}
& 73.5 & \textbf{71.5} & \underline{60.9} & \underline{57.1} & \underline{65.8} \\
\bottomrule
\end{tabular}
\end{table}

\section{Analysis of Process and Outcome Supervision}
\label{app:rsp_supervision_analysis}

We analyze how process annotations and outcome labels jointly supervise break and repair in \algo{}. We first relate the fit of adjacent annotated states to the corresponding head predictions. We then derive how a final outcome changes the relative support for intermediate states and how this conditioning enters the training gradients.

We retain the notation of Section~\ref{method} and omit the trajectory index $i$. For a fixed question and reasoning trajectory $(x,\mathbf{s})$, state propagation gives
\begin{equation}
p_t^{\mathrm{G}}
=
p_{t-1}^{\mathrm{G}}(1-\alpha_t)
+
p_{t-1}^{\mathrm{B}}\beta_t,
\qquad
p_t^{\mathrm{B}}=1-p_t^{\mathrm{G}}.
\label{eq:rsp_analysis_propagation}
\end{equation}
All probabilities below refer to reasoning states along this observed trajectory.

\subsection{Process Supervision of Break and Repair}
\label{app:rsp_local_grounding}

For an annotated step $t\in\mathcal{A}$, the individual state loss is
\begin{equation}
\ell_t^{\mathrm{step}}
=
-y_t^{\mathrm{step}}\log p_t^{\mathrm{G}}
-
(1-y_t^{\mathrm{step}})\log p_t^{\mathrm{B}}.
\label{eq:rsp_analysis_state_loss}
\end{equation}
When $p_{t-1}^{\mathrm{G}}=1$, propagation reduces to
$p_t^{\mathrm{G}}=1-\alpha_t$, so this loss is binary cross-entropy for $\alpha_t$ with target $1-y_t^{\mathrm{step}}$. When
$p_{t-1}^{\mathrm{B}}=1$, it instead reduces to binary cross-entropy for $\beta_t$ with target $y_t^{\mathrm{step}}$. Thus, fitting the current state supervises break after a valid prefix and repair after an invalid prefix. The same relation holds approximately when the preceding state is not deterministic.

\paragraph{Proposition 1 (Adjacent-state supervision).}
Suppose $t-1,t\in\mathcal{A}$, and let $a,b\in\mathcal{Z}$ denote the states specified by their process labels, with $1$ mapped to $\mathrm{G}$ and $0$ to $\mathrm{B}$. If
\begin{equation}
p_{t-1}^{a}\geq 1-\epsilon,
\qquad
p_t^{b}\geq 1-\delta,
\qquad
0\leq\epsilon,\delta<1,
\label{eq:rsp_analysis_local_conditions}
\end{equation}
then
\begin{equation}
\begin{aligned}
y_{t-1}^{\mathrm{step}}=1
&\quad\Longrightarrow\quad
\left|\alpha_t-(1-y_t^{\mathrm{step}})\right|
\leq \frac{\delta}{1-\epsilon},\\
y_{t-1}^{\mathrm{step}}=0
&\quad\Longrightarrow\quad
\left|\beta_t-y_t^{\mathrm{step}}\right|
\leq \frac{\delta}{1-\epsilon}.
\end{aligned}
\label{eq:rsp_analysis_head_bounds}
\end{equation}

\paragraph{Proof.}
Let $\bar b$ be the state other than $b$. Since all entries of
$\mathbf{M}_t$ are nonnegative,
\begin{equation}
\delta
\geq p_t^{\bar b}
\geq p_{t-1}^{a}\mathbf{M}_t[a,\bar b]
\geq (1-\epsilon)\bigl(1-\mathbf{M}_t[a,b]\bigr).
\label{eq:rsp_analysis_local_proof}
\end{equation}
Hence $1-\mathbf{M}_t[a,b]\leq\delta/(1-\epsilon)$. Substituting the appropriate entry of $\mathbf{M}_t$ gives both inequalities in
\eqref{eq:rsp_analysis_head_bounds}.
For example, when an invalid prefix is followed by a valid one and both states are well fitted, the bound requires $\beta_t$ to be close to one. The result applies to the prediction conditioned on the annotated preceding state. It establishes how process supervision constrains the meaning of break and repair through state propagation, without requiring separate
labels for the two heads.

\subsection{Outcome Evidence for Intermediate Reasoning States}
\label{app:rsp_outcome_conditioning}

For an outcome-annotated trajectory, let $c=\mathrm{G}$ when
$y^{\mathrm{out}}=1$ and $c=\mathrm{B}$ otherwise, following the terminal target used by $\mathcal{L}_{\mathrm{out}}$. Expanding the propagation product gives
\begin{equation}
Z_{\mathrm{out}}
\coloneqq p_T^{c}
=
\sum_{z_{1:T-1}\in\mathcal{Z}^{T-1}}
\prod_{j=1}^{T}\mathbf{M}_j[z_{j-1},z_j],
\label{eq:rsp_analysis_outcome_marginal}
\end{equation}
where $z_0=\mathrm{G}$ and $z_T=c$ are fixed. The individual outcome loss is $\ell^{\mathrm{out}}=-\log Z_{\mathrm{out}}$. The sum includes every possible sequence of intermediate states ending in $c$. A correct outcome can therefore be explained by reasoning that remains valid throughout or by reasoning that contains an error followed by recovery.

To compare these explanations at position $t$, define
\begin{equation}
h_t(a)
\coloneqq
\bigl(\mathbf{M}_{t+1}\cdots\mathbf{M}_T\bigr)[a,c],
\qquad a\in\mathcal{Z},
\label{eq:rsp_analysis_suffix_likelihood}
\end{equation}
with the empty product equal to the identity matrix. The quantity $h_t(a)$ is the probability of reaching the observed terminal state from state $a$ through the remaining reasoning steps. It satisfies
\begin{equation}
h_T(a)=\mathbf{1}\{a=c\},
\qquad
h_{t-1}(a)=\sum_{b\in\mathcal{Z}}\mathbf{M}_t[a,b]h_t(b).
\label{eq:rsp_analysis_backward}
\end{equation}

The joint probability of $z_t=a$ and $z_T=c$ is $p_t^{a}h_t(a)$.
Conditioning on the terminal label therefore gives
\begin{equation}
\gamma_t^{a}
\coloneqq
p_{\theta}(z_t=a\mid z_T=c,x,\mathbf{s})
=
\frac{p_t^{a}h_t(a)}{Z_{\mathrm{out}}},
\qquad
Z_{\mathrm{out}}=\sum_{a\in\mathcal{Z}}p_t^{a}h_t(a).
\label{eq:rsp_analysis_state_posterior}
\end{equation}
This expression combines the model's prefix evaluation $p_t^{a}$ with the likelihood of the final result under that state, $h_t(a)$. In particular,
\begin{equation}
\gamma_t^{\mathrm{G}}-p_t^{\mathrm{G}}
=
\frac{
p_t^{\mathrm{G}}p_t^{\mathrm{B}}
\bigl(h_t(\mathrm{G})-h_t(\mathrm{B})\bigr)
}{Z_{\mathrm{out}}}.
\label{eq:rsp_analysis_posterior_shift}
\end{equation}
The outcome \textbf{increases the relative support for whichever intermediate state makes the observed final result more likely under the remaining reasoning}. It leaves the state probabilities unchanged when the two explanations have equal likelihood. Thus, the terminal observation can support or revise a prefix prediction rather than simply reinforce its existing preference. These posterior probabilities describe conditioning at fixed model parameters, and the process score remains $p_t^{\mathrm{G}}$, computed from the prefix alone.

\subsection{Learning Break and Repair from the Final Outcome}
\label{app:rsp_outcome_gradients}

The posterior reweighting above also determines the gradients of the outcome loss. Let $\eta_t^{\mathrm{br}}$ and $\eta_t^{\mathrm{rp}}$ be the head logits, with $\alpha_t=\sigma(\eta_t^{\mathrm{br}})$ and
$\beta_t=\sigma(\eta_t^{\mathrm{rp}})$. The outcome-conditioned targets for break and repair are
\begin{equation}
\widetilde\alpha_t
=
\frac{\alpha_t h_t(\mathrm{B})}{h_{t-1}(\mathrm{G})},
\qquad
\widetilde\beta_t
=
\frac{\beta_t h_t(\mathrm{G})}{h_{t-1}(\mathrm{B})}.
\label{eq:rsp_analysis_soft_targets}
\end{equation}
For finite logits, both denominators are positive. Where the corresponding preceding state has positive probability, these quantities are respectively
$p_{\theta}(z_t=\mathrm{B}\mid z_{t-1}=\mathrm{G},z_T=c,x,\mathbf{s})$
and
$p_{\theta}(z_t=\mathrm{G}\mid z_{t-1}=\mathrm{B},z_T=c,x,\mathbf{s})$.

\paragraph{Proposition 2 (Outcome supervision of the two heads).}
The local derivatives of the outcome loss are
\begin{equation}
\begin{aligned}
\frac{\partial\ell^{\mathrm{out}}}{\partial\eta_t^{\mathrm{br}}}
&=
\gamma_{t-1}^{\mathrm{G}}
\bigl(\alpha_t-\widetilde\alpha_t\bigr),\\
\frac{\partial\ell^{\mathrm{out}}}{\partial\eta_t^{\mathrm{rp}}}
&=
\gamma_{t-1}^{\mathrm{B}}
\bigl(\beta_t-\widetilde\beta_t\bigr).
\end{aligned}
\label{eq:rsp_analysis_logit_gradients}
\end{equation}

\paragraph{Proof.}
Holding the other head outputs fixed, write
\begin{equation}
\begin{aligned}
Z_{\mathrm{out}}
={}&p_{t-1}^{\mathrm{G}}
\bigl((1-\alpha_t)h_t(\mathrm{G})+\alpha_t h_t(\mathrm{B})\bigr)\\
&+p_{t-1}^{\mathrm{B}}
\bigl(\beta_t h_t(\mathrm{G})+(1-\beta_t)h_t(\mathrm{B})\bigr).
\end{aligned}
\label{eq:rsp_analysis_local_likelihood}
\end{equation}
Differentiating $-\log Z_{\mathrm{out}}$ through the sigmoid gives
\begin{equation}
\begin{aligned}
\frac{\partial\ell^{\mathrm{out}}}{\partial\eta_t^{\mathrm{br}}}
&=
\frac{p_{t-1}^{\mathrm{G}}\alpha_t(1-\alpha_t)}{Z_{\mathrm{out}}}
\bigl(h_t(\mathrm{G})-h_t(\mathrm{B})\bigr),\\
\frac{\partial\ell^{\mathrm{out}}}{\partial\eta_t^{\mathrm{rp}}}
&=
-\frac{p_{t-1}^{\mathrm{B}}\beta_t(1-\beta_t)}{Z_{\mathrm{out}}}
\bigl(h_t(\mathrm{G})-h_t(\mathrm{B})\bigr).
\end{aligned}
\label{eq:rsp_analysis_direct_gradients}
\end{equation}
Substituting \eqref{eq:rsp_analysis_backward},
\eqref{eq:rsp_analysis_state_posterior}, and
\eqref{eq:rsp_analysis_soft_targets} yields
\eqref{eq:rsp_analysis_logit_gradients}.

\Eqref{eq:rsp_analysis_logit_gradients} has the form of a weighted binary cross-entropy gradient. Each head is compared with its outcome-conditioned target and weighted by the posterior probability of the relevant preceding state. For example, a repair target larger than $\beta_t$ gives a negative derivative for the repair logit, encouraging a higher repair prediction. If $h_t(\mathrm{G})=h_t(\mathrm{B})>0$, the two
targets equal the current predictions and both local derivatives vanish. The final result therefore supplies a learning signal according to how a step relates to the subsequent reasoning, rather than imposing the same validity target at every position.

These equalities describe the gradient of the existing outcome objective. In the weighted cross-entropy interpretation, the posterior quantities are evaluated at the current parameters and treated as fixed targets and weights. Automatic differentiation of $\ell^{\mathrm{out}}$ produces the
same local derivatives without an explicit posterior-labeling stage.


\paragraph{Implication for joint supervision.}
Process annotations provide direct evidence about intermediate validity through \eqref{eq:rsp_analysis_head_bounds}. On outcome-only data, the same model supplies the prefix probabilities used in \eqref{eq:rsp_analysis_state_posterior}, while the terminal label reweights these predictions through the remaining reasoning. The two losses consequently train the same break and repair heads using different levels of annotation.

A useful instance is a correct outcome following a predicted invalid prefix. If $y^{\mathrm{out}}=1$,
$p_k^{\mathrm{B}}\geq1-\epsilon$, and
$p_T^{\mathrm{G}}\geq1-\delta$, with $0\leq\epsilon,\delta<1$, then
\begin{equation}
h_k(\mathrm{B})
\geq
1-\frac{\delta}{1-\epsilon}.
\label{eq:rsp_analysis_recovery_bound}
\end{equation}
This follows from
$p_T^{\mathrm{B}}\geq p_k^{\mathrm{B}}(1-h_k(\mathrm{B}))$.
Conditional on retaining the invalid prefix prediction, fitting the correct outcome requires the remaining steps to provide sufficient recovery. The outcome constrains their combined effect while leaving the location of recovery to be learned from the reasoning context. This connects locally supervised validity judgments with terminal feedback on trajectories that
lack process annotations. Section~\ref{exp:rsp_analysis} and
Table~\ref{tab:ablation} evaluate the learned break and repair scores and the empirical contribution of the two supervision sources.

\subsection{Strength of State Propagation}
\label{app:state_propagation_strength}

The analysis above shows that outcome supervision reaches intermediate
transitions through the propagated reasoning states. We further examine
the strength of this propagation in the learned model. From
\eqref{eq:rsp_analysis_propagation}, using
$p_{t-1}^{\mathrm{B}}=1-p_{t-1}^{\mathrm{G}}$, we can write
\begin{equation}
p_t^{\mathrm{G}}
=
\beta_t
+
\kappa_t p_{t-1}^{\mathrm{G}},
\qquad
\kappa_t
\coloneqq
1-\alpha_t-\beta_t.
\label{eq:rsp_kappa}
\end{equation}
Therefore,
\begin{equation}
\frac{\partial p_t^{\mathrm{G}}}
     {\partial p_{t-1}^{\mathrm{G}}}
=
\kappa_t.
\label{eq:rsp_kappa_local}
\end{equation}
The coefficient $\kappa_t$ directly measures how strongly the propagated
state after step $t$ depends on the preceding propagated state. Across
multiple steps,
\begin{equation}
\frac{\partial p_t^{\mathrm{G}}}
     {\partial p_s^{\mathrm{G}}}
=
\prod_{j=s+1}^{t}\kappa_j,
\qquad s<t.
\label{eq:rsp_kappa_forward}
\end{equation}
Thus, smaller $|\kappa_t|$ weakens the influence of earlier reasoning
states, whereas larger $|\kappa_t|$ preserves more of the preceding
state information.

The same coefficient also determines how outcome supervision is
transmitted backward through state propagation. Recall that
$h_t(a)$ denotes the probability of reaching the observed terminal
state from state $a$. From the backward recursion in
\eqref{eq:rsp_analysis_backward}, the difference between the two
possible current states satisfies
\begin{equation}
h_{t-1}(\mathrm{G})-h_{t-1}(\mathrm{B})
=
\kappa_t
\bigl(
h_t(\mathrm{G})-h_t(\mathrm{B})
\bigr).
\label{eq:rsp_kappa_backward}
\end{equation}
Consequently,
\begin{equation}
\left|
h_t(\mathrm{G})-h_t(\mathrm{B})
\right|
=
\prod_{j=t+1}^{T}
|\kappa_j|,
\label{eq:rsp_kappa_outcome}
\end{equation}
where the terminal difference has magnitude one. Since
$h_t(\mathrm{G})-h_t(\mathrm{B})$ appears directly in the outcome
gradients in \eqref{eq:rsp_analysis_direct_gradients}, $\kappa_t$
has a dual role: it controls both the forward dependence of later
states on earlier states and the backward propagation of outcome
supervision to earlier transitions. Importantly, this formulation does not require outcome supervision to propagate equally strongly to all preceding transitions. Its influence is modulated by the learned intervening state transitions, so that earlier states whose effect is weakly preserved receive correspondingly weaker terminal evidence.

\begin{figure*}[t]
    \centering

    \begin{minipage}[t]{0.49\textwidth}
        \centering
        \includegraphics[height=5.0cm]{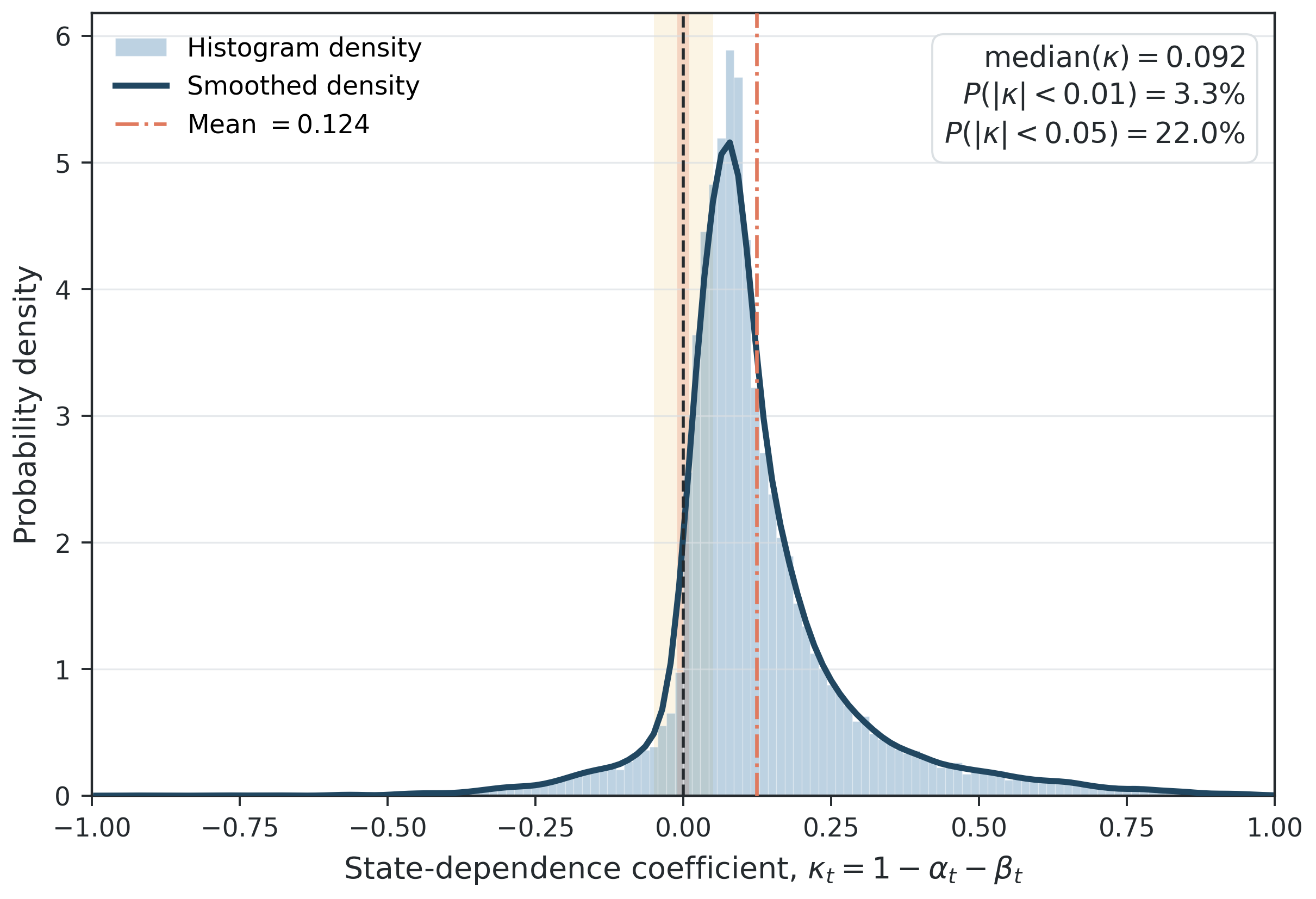}
        \vspace{-2mm}
        \centerline{\small (a) Distribution of $\kappa_t$}
    \end{minipage}
    \hfill
    \begin{minipage}[t]{0.49\textwidth}
        \centering
        \includegraphics[height=5.0cm]{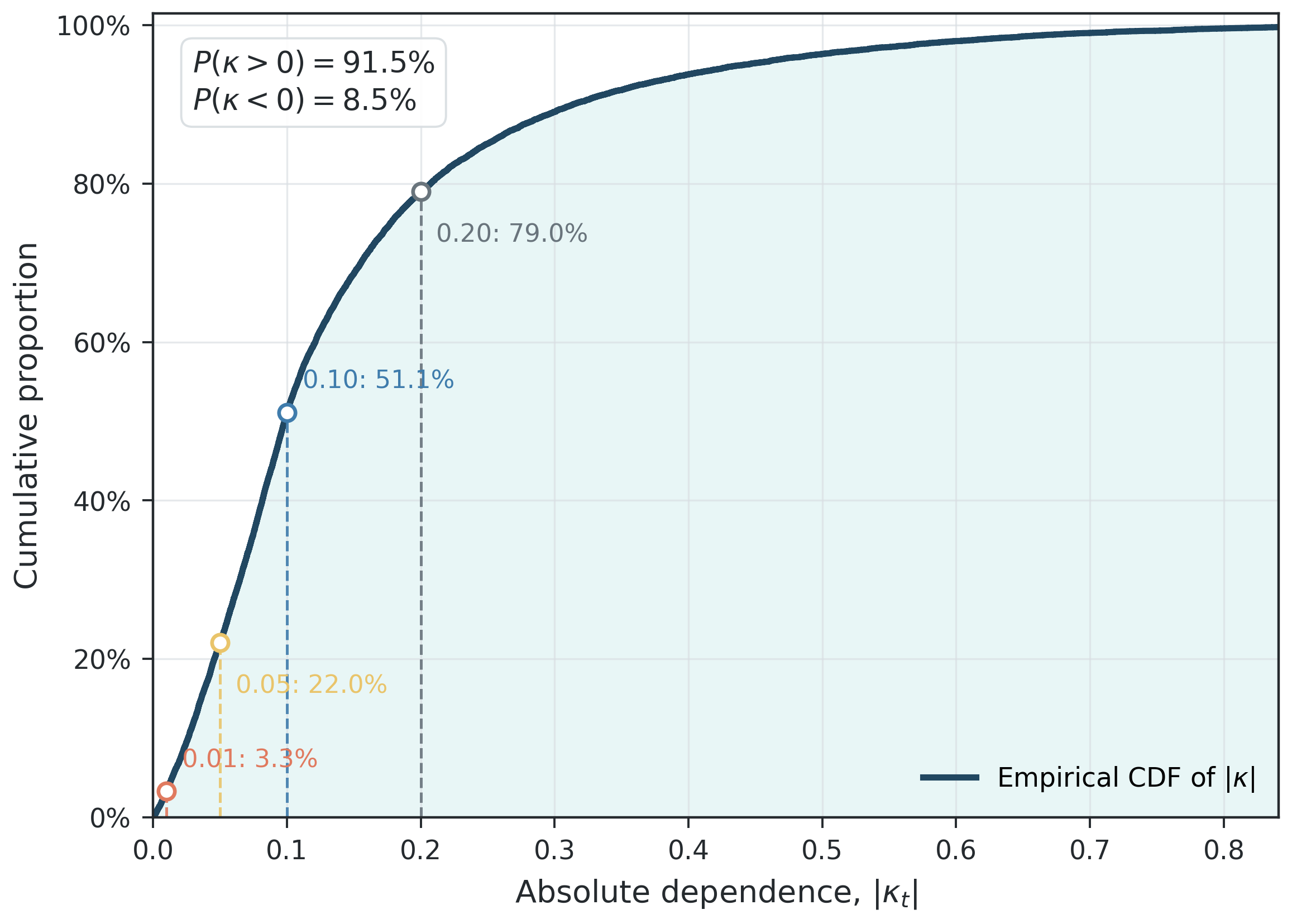}
        \vspace{-2mm}
        \centerline{\small (b) Empirical cumulative distribution of $|\kappa_t|$}
    \end{minipage}

    \caption{
    (a) Distribution of $\kappa_t=1-\alpha_t-\beta_t$.
    The signed value indicates the direction and strength of the dependence
    of the updated propagated state on the preceding state.
    (b) Empirical cumulative distribution of $|\kappa_t|$.
    Smaller $|\kappa_t|$ indicates weaker dependence of the updated
    propagated state on the preceding propagated state.
    }
    \label{fig:kappa_analysis}
\end{figure*}

\paragraph{Empirical Analysis}
Figure~\ref{fig:kappa_analysis}(a) shows that the learned state dependence
is predominantly positive. The median value of $\kappa_t$ is $0.092$ and
the mean is $0.124$. Moreover, $91.5\%$ of the transitions have
$\kappa_t>0$, while only $8.5\%$ have $\kappa_t<0$. Therefore, for the
large majority of reasoning steps, a higher probability of being in the
valid state before the step continues to increase the propagated
valid state probability after the step. The distribution also contains
a broad positive tail, showing that a subset of transitions retains
substantially stronger dependence on preceding states.

Figure~\ref{fig:kappa_analysis}(b) further shows that the propagation
strength varies considerably across reasoning steps. Only $3.3\%$ of
transitions satisfy $|\kappa_t|<0.01$, indicating that transitions that
almost completely remove dependence on the preceding propagated state
are relatively uncommon. Meanwhile, $22.0\%$ of transitions have
$|\kappa_t|<0.05$, $51.1\%$ have $|\kappa_t|<0.1$, and $79.0\%$ have
$|\kappa_t|<0.2$. Thus, \algo{} does not maintain the same propagation
strength at every reasoning step. Instead, some transitions substantially
reduce the influence of preceding states, while others preserve stronger
dependence across successive steps.

These results suggest that reasoning state propagation is selective
rather than uniform. This behavior is consistent with the formulation
of \algo{}: later reasoning may preserve an earlier reasoning state,
introduce an error, or recover from an earlier error, and these cases
need not retain the same amount of dependence on the preceding state.
The same learned transition structure also determines how outcome
supervision is propagated backward. Therefore, \algo{} allows both
state dependence and outcome supervision to vary across the reasoning
trajectory according to the learned state transitions, rather than
requiring equally strong propagation at every step.

\section{Scaling Analysis}
\label{app:scaling}

\begin{figure*}[t]
    \centering
    \includegraphics[width=0.95\textwidth]{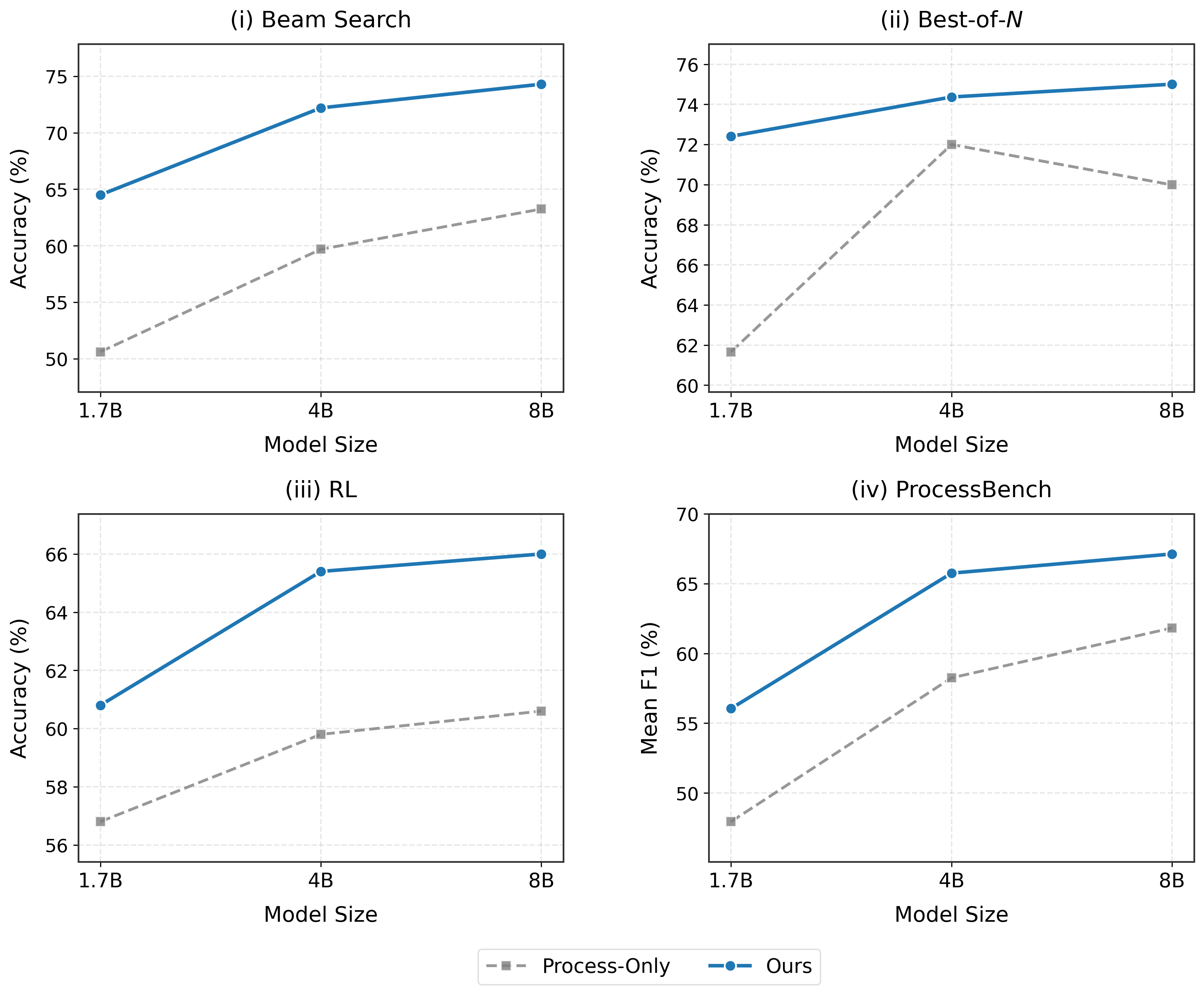}
    \caption{
    Performance with different PRM backbone sizes.
    We compare \algo{} with Process-Only on beam search, Best-of-$N$ selection,
    reinforcement learning, and ProcessBench.
    }
    \label{fig:model_scaling}
\end{figure*}

We further examine how \algo{} behaves under different model capacities and amounts of training supervision. Unless otherwise specified, all training and evaluation settings follow the main experiments. \emph{Process-Only} denotes the corresponding model trained with process supervision alone.


\paragraph{Scaling with Model Size.}
We first study whether the benefit of \algo{} persists as the PRM backbone scales. We vary the backbone size of the Qwen3 series models from 1.7B to 8B while keeping the training data and other settings fixed. As shown in Figure~\ref{fig:model_scaling}, \algo{} consistently improves as the model size increases across all four evaluation settings. Compared with Process-Only, \algo{} maintains a clear advantage across all model sizes. These results indicate that the benefit of incorporating outcome supervision is maintained as the PRM capacity increases.

\begin{figure*}[t]
    \centering
    \includegraphics[width=0.95\textwidth]{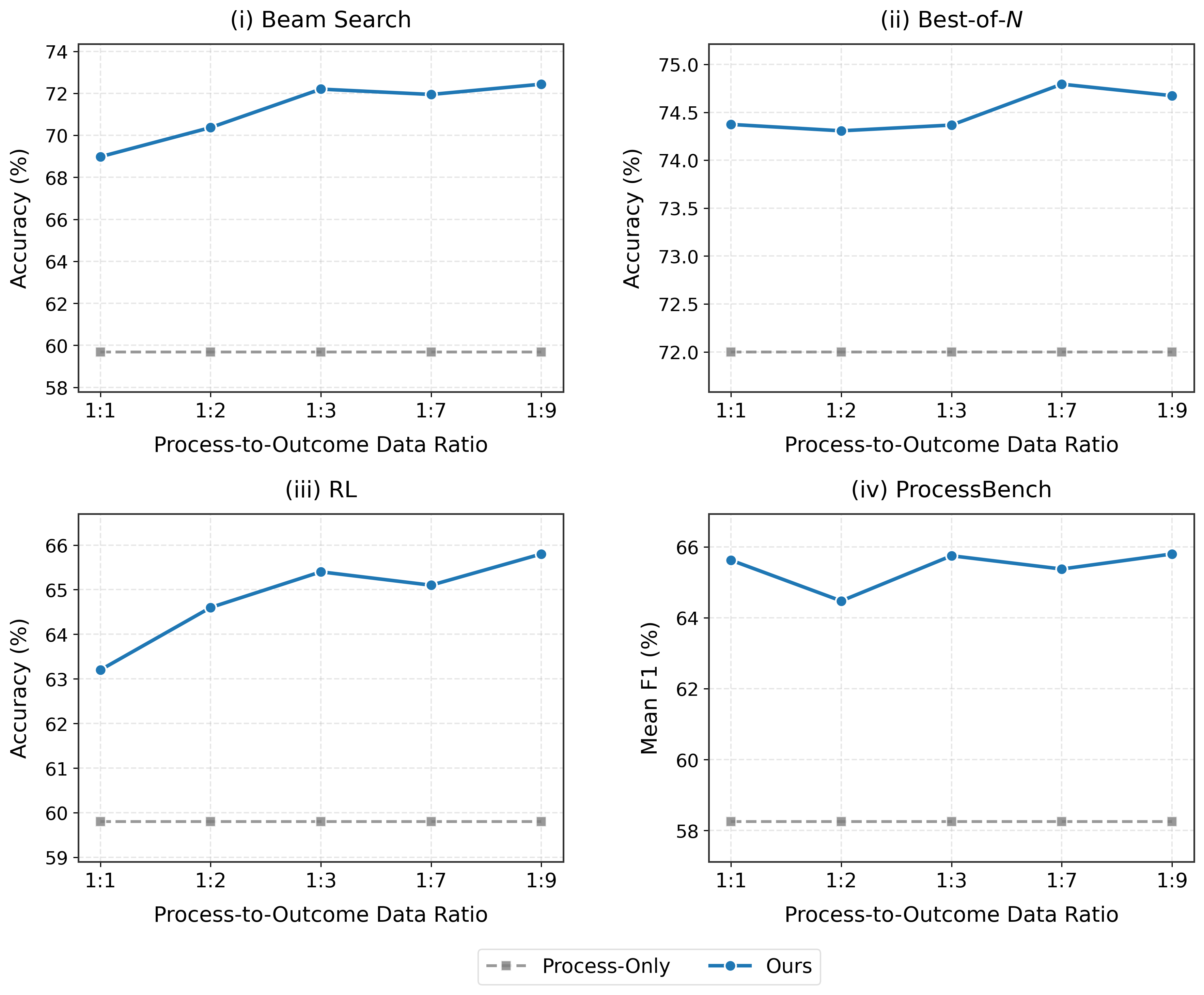}
    \caption{
    Performance with different amounts of outcome supervision. The horizontal axis denotes the process-to-outcome data ratio within each
    training batch. For example, $1{:}3$ indicates that the numbers of process-annotated and outcome-annotated examples are mixed at a ratio of
    $1{:}3$ in a batch. The process supervision is kept fixed while the amount of outcome supervision is varied.
    }
    \label{fig:outcome_scaling}
\end{figure*}

\paragraph{Scaling with Outcome Supervision.}
We further study how \algo{} scales with the amount of outcome supervision by varying the process-to-outcome data ratio in each training batch from $1{:}1$
to $1{:}9$, while keeping the process supervision fixed. As shown in Figure~\ref{fig:outcome_scaling}, increasing outcome supervision provides substantial improvements in beam search and reinforcement learning. Beam search accuracy increases from approximately 69\% to above 72\%, while reinforcement learning improves from about 63.2\% to 65.8\%. In contrast, Best-of-$N$ selection and ProcessBench remain relatively stable across different ratios. These results show that \algo{} can effectively benefit from additional outcome supervision, particularly for reasoning search and policy optimization.

\section{Training Data and Evaluation Overlap}
\label{app:data_overlap}

We further clarify the relation between the training data used in our experiments and the evaluation benchmarks.

\paragraph{Process Supervision.}
For process supervision, we use a deduplicated version of PRM800K~\citep{lightman2024let} before selecting the 20\% subset used in our main experiments.
Specifically, duplicated full reasoning trajectories are removed during preprocessing.
The deduplication is performed at the trajectory level.
Distinct trajectories that share partial reasoning prefixes are retained, since they correspond to different annotated reasoning trajectories.
PRM800K uses the MATH split released by \citet{lightman2024let}.
In this split, 4,500 problems from the original MATH test set are included in the training set, while the remaining 500 problems are held out for evaluation.
These held-out problems form MATH500 and are not included in the PRM800K training data used for process supervision. Therefore, our MATH500 evaluation is separated from the PRM800K process training data by construction.

\paragraph{Outcome Supervision.}
For outcome supervision, we use the released AceMath-RM~\citep{liu2025acemathadvancingfrontiermath} training data and convert it into outcome-annotated trajectories following Appendix~\ref{app:outcome_data_construction}.
Our preprocessing only filters examples and converts each retained response into reasoning steps with a trajectory-level outcome label.
It does not introduce additional questions from any evaluation benchmark.
AceMath-RM is constructed from the math training data of AceMath, where test-data filtering is applied before model training.


\paragraph{Reinforcement Learning.}
The verifier-gated update used in our reinforcement learning experiments follows VeriGate~\citep{agrawal2026verigate}.
For RL training data, we use an 8K subset of the public DAPO-MATH-17K dataset~\citep{yu2026dapo}, following the setting described in Section~\ref{exp:rl} and Appendix~\ref{app:rl_details}.
DAPO-MATH-17K is released as a curated and processed math training dataset.
We sample the 8K training prompts only from this released training set.
All methods in Table~\ref{tab:rl_results} use exactly the same 8K training prompts, together with the same policy initialization and optimization budget.

Overall, the process, outcome, and reinforcement learning experiments use their corresponding public training data.
Our preprocessing does not add examples from the evaluation benchmarks to these training sets.
In particular, MATH500 is explicitly held out from the PRM800K process training split.
The same training data are used across the compared methods under each experimental setting.

\section{Qualitative Examples}
\label{app:examples}

We provide qualitative examples to illustrate how \algo{} propagates reasoning validity across different reasoning patterns. Each step reports the propagated process score together with the predicted \emph{break} and \emph{repair} probabilities. Examples illustrate representative state patterns, including validity preservation, error introduction and persistence, and error recovery. 


\begin{prmexample}{Example 1}

\textbf{Question:}

How many four-digit positive integers contain at least one $2$ or $3$
as a digit?

\vspace{0.55em}

\textbf{LLM Response:}


\CorrectStepHeader
    {1}
    {0.992}
    {0.008}
    {0.909}

To determine how many four-digit positive integers contain at least
one $2$ or $3$ as a digit, we can use the principle of complementary
counting. This involves first finding the total number of four-digit
positive integers and then subtracting the number of four-digit
positive integers that do not contain any 2s or 3s as digits.


\CorrectStepHeader
    {2}
    {0.997}
    {0.002}
    {0.910}

Calculate the total number of four-digit positive integers.

A four-digit positive integer ranges from $1000$ to $9999$.
Therefore, the total number of four-digit positive integers is:
\[
9999 - 1000 + 1 = 9000.
\]


\CorrectStepHeader
    {3}
    {0.995}
    {0.005}
    {0.905}

Calculate the number of four-digit positive integers that do not
contain any 2s or 3s as digits.

For a four-digit number, the digits can be chosen as follows:

\begin{itemize}
    \item The thousands place can be any digit from $1$ to $9$
    except $2$ and $3$, so there are $7$ choices
    $(1,4,5,6,7,8,9)$.

    \item The hundreds, tens, and units places can each be any digit
    from $0$ to $9$ except $2$ and $3$, so there are $8$ choices
    for each of these places
    $(0,1,4,5,6,7,8,9)$.
\end{itemize}


\CorrectStepHeader
    {4}
    {0.996}
    {0.003}
    {0.887}

Thus, the total number of four-digit positive integers that do not
contain any 2s or 3s as digits is:
\[
7 \times 8 \times 8 \times 8
=
7 \times 512
=
3584.
\]


\CorrectStepHeader
    {5}
    {0.997}
    {0.002}
    {0.853}

Calculate the number of four-digit positive integers that contain
at least one $2$ or $3$ as a digit.

Using the principle of complementary counting, we subtract the number
of four-digit positive integers that do not contain any 2s or 3s
from the total number of four-digit positive integers:
\[
9000 - 3584 = 5416.
\]


\CorrectStepHeader
    {6}
    {1.000}
    {0.000}
    {0.835}

Therefore, the number of four-digit positive integers that contain
at least one $2$ or $3$ as a digit is:
\[
\boxed{5416}.
\]

\end{prmexample}

\paragraph{$\mathrm{G}\!\rightarrow\!\mathrm{G}$: Validity Preservation.}
Example~1 shows a fully valid reasoning trajectory. The break probability remains close to zero throughout the solution, and the propagated process score stays near one, correctly preserving the valid reasoning state across successive steps.

\begin{prmexample}{Example 2}

\textbf{Question:}

Given that the function $f(x)$ defined on $\mathbb{R}$ is an odd
function and satisfies $f(1+x)=f(3+x)$. When $0\leq x\leq 1$,
$f(x)=x^{3}-x$. Find
\[
f\left(\frac{11}{2}\right)+f(6).
\]

\vspace{0.55em}

\textbf{LLM Response:}


\CorrectStepHeader
    {1}
    {0.991}
    {0.009}
    {0.973}

To find $f\left(\frac{11}{2}\right)+f(6)$, we need to use the
properties of the function $f$. The function $f$ is an odd function,
which means $f(-x)=-f(x)$, and it satisfies the periodicity condition
$f(1+x)=f(3+x)$. This implies that $f(x)=f(x+2)$, so $f$ is periodic
with period $2$.


\CorrectStepHeader
    {2}
    {0.992}
    {0.008}
    {0.998}

First, let's find $f(6)$. Since $f$ is periodic with period $2$,
we have:
\[
f(6)=f(6-2\cdot3)=f(0).
\]

Given that $f(x)=x^3-x$ for $0\leq x\leq1$, we substitute $x=0$:
\[
f(0)=0^3-0=0.
\]

So, $f(6)=0$.


\IncorrectStepHeader
    {3}
    {0.104}
    {0.895}
    {0.011}

Next, let's find $f\left(\frac{11}{2}\right)$. Using the periodicity
of $f$, we have:
\[
f\left(\frac{11}{2}\right)
=
f\left(\frac{11}{2}-2\cdot2\right)
=
f\left(\frac{11}{2}-4\right)
=
f\left(\frac{3}{2}\right).
\]

Since $f$ is odd, we can write:
\[
f\left(\frac{3}{2}\right)
=
f\left(1+\frac{1}{2}\right)
=
f\left(3+\frac{1}{2}-2\right)
=
f\left(1+\frac{1}{2}\right)
=
f\left(\frac{1}{2}\right).
\]

Given that $f(x)=x^3-x$ for $0\leq x\leq1$, we substitute
$x=\frac{1}{2}$:
\[
f\left(\frac{1}{2}\right)
=
\left(\frac{1}{2}\right)^3-\frac{1}{2}
=
\frac{1}{8}-\frac{1}{2}
=
\frac{1}{8}-\frac{4}{8}
=
-\frac{3}{8}.
\]

So,
\[
f\left(\frac{11}{2}\right)=-\frac{3}{8}.
\]


\IncorrectStepHeader
    {4}
    {0.018}
    {0.870}
    {0.005}

Finally, we add $f\left(\frac{11}{2}\right)$ and $f(6)$:
\[
f\left(\frac{11}{2}\right)+f(6)
=
-\frac{3}{8}+0
=
-\frac{3}{8}.
\]

Thus, the answer is:
\[
\boxed{-\frac{3}{8}}.
\]

\end{prmexample}

\paragraph{$\mathrm{G}\!\rightarrow\!\mathrm{B}$ and
$\mathrm{B}\!\rightarrow\!\mathrm{B}$: Error Introduction and Persistence.}
In Example~2, the reasoning is valid through Step~2, but Step~3 incorrectly reduces $f(3/2)$ to $f(1/2)$. The high break probability sharply lowers the process score, while the subsequent step continues from the incorrect result with a very low repair probability, preserving the invalid reasoning state.


\begin{prmexample}{Example 3}

\textbf{Question:}

Rationalize the denominator of
\[
\frac{7}{\sqrt{75}}.
\]

\vspace{0.55em}

\textbf{LLM Response:}


\CorrectStepHeader
    {1}
    {0.982}
    {0.018}
    {0.762}

To rationalize the denominator of the expression
$\frac{7}{\sqrt{75}}$, we need to eliminate the square root from
the denominator. We can do this by multiplying both the numerator
and the denominator by $\sqrt{75}$.

Here are the steps:

Start with the original expression:
\[
\frac{7}{\sqrt{75}}.
\]


\CorrectStepHeader
    {2}
    {0.995}
    {0.004}
    {0.941}

Multiply both the numerator and the denominator by $\sqrt{75}$:
\[
\frac{7}{\sqrt{75}}
\cdot
\frac{\sqrt{75}}{\sqrt{75}}.
\]


\CorrectStepHeader
    {3}
    {0.987}
    {0.012}
    {0.738}

Perform the multiplication in the numerator and the denominator:
\[
\frac{7\cdot\sqrt{75}}
     {\sqrt{75}\cdot\sqrt{75}}
=
\frac{7\sqrt{75}}{75}.
\]


\IncorrectStepHeader
    {4}
    {0.399}
    {0.600}
    {0.358}

Simplify the expression by canceling the common factor of $7$
in the numerator and the denominator:
\[
\frac{7\sqrt{75}}{75}.
\]


\CorrectStepHeader
    {5}
    {0.802}
    {0.018}
    {0.682}

Simplify $\sqrt{75}$. We know that $75=25\cdot3$ and
$\sqrt{25}=5$, so:
\[
\sqrt{75}
=
\sqrt{25\cdot3}
=
5\sqrt{3}.
\]


\CorrectStepHeader
    {6}
    {0.924}
    {0.009}
    {0.651}

Substitute $\sqrt{75}$ with $5\sqrt{3}$ in the expression:
\[
\frac{7\cdot5\sqrt{3}}{75}
=
\frac{35\sqrt{3}}{75}.
\]


\CorrectStepHeader
    {7}
    {0.925}
    {0.044}
    {0.556}

Simplify the fraction $\frac{35}{75}$. The greatest common divisor
of $35$ and $75$ is $5$, so:
\[
\frac{35}{75}
=
\frac{35\div5}{75\div5}
=
\frac{7}{15}.
\]


\CorrectStepHeader
    {8}
    {0.972}
    {0.024}
    {0.929}

Therefore, the expression simplifies to:
\[
\frac{7\sqrt{3}}{15}.
\]


\CorrectStepHeader
    {9}
    {0.999}
    {0.000}
    {0.972}

The rationalized form of the expression is
\[
\boxed{\frac{7\sqrt{3}}{15}}.
\]

\end{prmexample}

\paragraph{$\mathrm{B}\!\rightarrow\!\mathrm{G}$: Error Recovery.}
Example~3 illustrates recovery after an intermediate error. Step~4 introduces an invalid reasoning step and reduces the process score, whereas the subsequent valid derivation returns to a correct computation. Correspondingly, the repair probability increases and the propagated process score gradually recovers.


\begin{prmexample}{Example 4}

\textbf{Question:}

Calculate the sum of all integral values of $c$ with $c \leq 20$
for which the equation
\[
y=x^2-9x-c
\]
produces two rational roots.

\vspace{0.55em}

\textbf{LLM Response:}


\CorrectStepHeader
    {1}
    {0.955}
    {0.045}
    {0.890}

To determine the sum of all integral values of $c$ with $c\leq20$
for which the quadratic equation $y=x^2-9x-c$ has two rational
roots, we start by using the discriminant. A quadratic equation
$ax^2+bx+c=0$ has two rational roots if and only if its
discriminant
\[
\Delta=b^2-4ac
\]
is a perfect square.


\CorrectStepHeader
    {2}
    {0.985}
    {0.012}
    {0.914}

For the given equation $x^2-9x-c=0$, the discriminant is
\[
\Delta=(-9)^2-4\cdot1\cdot(-c)=81+4c.
\]
We need $81+4c$ to be a perfect square. Let
\[
81+4c=k^2
\]
for some integer $k$. Then
\[
c=\frac{k^2-81}{4}.
\]
Since $c$ must be an integer and $81\equiv1\pmod 4$, we require
\[
k^2\equiv1\pmod 4.
\]
Therefore, $k$ must be an odd integer.


\IncorrectStepHeader
    {3}
    {0.431}
    {0.570}
    {0.500}

Next, we need to find the values of $k$ such that $c\leq20$.
Substituting $c=\frac{k^2-81}{4}$ into the inequality gives
\[
\frac{k^2-81}{4}\leq20
\quad\Rightarrow\quad
k^2\leq161.
\]
The largest odd integer $k$ that satisfies $k^2\leq161$ is $12$,
and the smallest odd integer is $-13$. Therefore, $k$ can take
any odd value from $-13$ to $13$.


\CorrectStepHeader
    {4}
    {0.890}
    {0.008}
    {0.812}

We now list the odd integers in this range and calculate the
corresponding values of $c$:
\[
\begin{aligned}
k=-13
&\quad\Rightarrow\quad
c=\frac{169-81}{4}=22
&&(\text{not valid since }c\leq20),\\
k=-11
&\quad\Rightarrow\quad
c=\frac{121-81}{4}=10,\\
k=-9
&\quad\Rightarrow\quad
c=\frac{81-81}{4}=0,\\
k=-7
&\quad\Rightarrow\quad
c=\frac{49-81}{4}=-8,\\
k=-5
&\quad\Rightarrow\quad
c=\frac{25-81}{4}=-14,\\
k=-3
&\quad\Rightarrow\quad
c=\frac{9-81}{4}=-18,\\
k=-1
&\quad\Rightarrow\quad
c=\frac{1-81}{4}=-20.
\end{aligned}
\]
The positive values of $k$ give the same values of $c$. In
particular, $k=13$ also gives $c=22$, which is not valid since
$c\leq20$. Therefore, the valid values of $c$ are
\[
-20,\,-18,\,-14,\,-8,\,0,\,10.
\]


\CorrectStepHeader
    {5}
    {0.978}
    {0.000}
    {0.805}

Summing these values, we get
\[
-20+(-18)+(-14)+(-8)+0+10=-50.
\]
Thus, the sum of all integral values of $c$ is
\[
\boxed{-50}.
\]

\end{prmexample}

\paragraph{$\mathrm{B}\!\rightarrow\!\mathrm{G}$: Error Recovery.}
Example~4 illustrates recovery after an intermediate error.
Step~3 incorrectly includes $\pm13$ among the possible values of
$k$, even though $13^2>161$. In Step~4, the response evaluates
these values, obtains $c=22$, and explicitly excludes them because
$c\leq20$. The derivation then returns to the correct set of values
and produces the correct final answer.

\end{document}